\documentclass{IEEEtran}
\usepackage[noadjust]{cite}
\ifCLASSINFOpdf
\else
\fi
\usepackage[T1]{fontenc} % optional enhanced font encoding
\usepackage[cmintegrals]{newtxmath}
\usepackage{bm}
\usepackage{array}
\usepackage{url}
\usepackage{cite}
\usepackage{amsmath,amsfonts}
\usepackage{algorithm}
\usepackage{algpseudocode}
\usepackage{graphicx}
\usepackage{textcomp}
\usepackage{xcolor}
\usepackage{balance}
\usepackage{wrapfig}
\usepackage{float}
\usepackage{subfig}
\usepackage{times}
\usepackage{caption}
\usepackage{makecell}
\usepackage{multirow}

\begin{document}

\title{MECKD: Deep Learning-Based Fall Detection in Multilayer Mobile Edge Computing With Knowledge Distillation}

\author{Wei-Lung Mao, Chun-Chi Wang,\\ Po-Heng Chou, \IEEEmembership{Member, IEEE}, Kai-Chun Liu, \IEEEmembership{Member, IEEE} and Yu Tsao, \IEEEmembership{Senior Member, IEEE}\\ \thanks{Manuscript received September xx; revised xx; accepted xx. Date of publication xx; date of current version xx. This work was supported in part by the Academia Sinica under Grant 235g Postdoctoral Scholar Program and ASGC-111-M0, and in part by the National Science and Technology Council (NSTC) of Taiwan under Grant 113-2926-I-001-502-G \emph{(Corresponding author: Po-Heng Chou)}.}
\thanks{Wei-Lung Mao and Chun-Chi Wang are with the Department of Electrical Engineering and Graduate School of Engineering Science and Technology, National Yunlin University of Science and Technology, Yunlin, 64002, Taiwan (e-mail: wlmao@yuntech.edu.tw, d11010202@gemail.yuntech.edu.tw). }
\thanks{Po-Heng Chou, Kai-Chun Liu are with Research Center for Information Technology Innovation (CITI), Academia Sinica, Taipei 11529, Taiwan (e-mail: d00942015@ntu.edu.tw, t22302856@gmail.com, yu.tsao@citi.sinica.edu.tw).}
\thanks{Yu Tsao is with the Research Center for Information Technology Innovation, Academia Sinica, Taipei 11529, Taiwan, and also with the Department of Electrical Engineering, Chung Yuan Christian University, Taoyuan 320, Taiwan (e-mail: yu.tsao@citi.sinica.edu.tw).}}

\maketitle
\begin{abstract}
The rising aging population has increased the importance of fall detection (FD) systems as an assistive technology, where deep learning techniques are widely applied to enhance accuracy. FD systems typically use edge devices (EDs) worn by individuals to collect real-time data, which are transmitted to a cloud center (CC) or processed locally. However, this architecture faces challenges such as a limited ED model size and data transmission latency to the CC. Mobile edge computing (MEC), which allows computations at MEC servers deployed between EDs and CC, has been explored to address these challenges. We propose a multilayer MEC (MLMEC) framework to balance accuracy and latency. The MLMEC splits the architecture into stations, each with a neural network model. If front-end equipment cannot detect falls reliably, data are transmitted to a station with more robust back-end computing. The knowledge distillation (KD) approach was employed to improve front-end detection accuracy by allowing high-power back-end stations to provide additional learning experiences, enhancing precision while reducing latency and processing loads. Simulation results demonstrate that the KD approach improved accuracy by 11.65\% on the SisFall dataset and 2.78\% on the FallAllD dataset. The MLMEC with KD also reduced the data latency rate by 54.15\% on the FallAllD dataset and 46.67\% on the SisFall dataset compared to the MLMEC without KD. In summary, the MLMEC FD system exhibits improved accuracy and reduced latency.
\end{abstract}


\IEEEpeerreviewmaketitle

\section{Introduction}
\label{sec:introduction}
\IEEEPARstart{F}{all}  are a major cause of severe injury and death among elderly individuals worldwide. According to the World Population Prospects 2022~\cite{re1_United_Nations}, there are 80 million people aged 65 and above, representing 10\% of the global population. Previous studies show that approximately half of fall victims are unable to get back up without assistance because of injuries or a lack of physical fitness and strength due to fall-related injuries or a lack of physical fitness and strength. Moreover, more than half of those injured face a high risk of death within six months~\cite{re2_Towards, re3_How_dangerous}. In the past two decades, wearable-based fall detection (FD) systems have been developed as an assistive technology~\cite{Review1_Add2,re4_Fall_classification, re5_Fall_detection_montoring, re6_Near-Fall}. The primary goal of FD systems is to automatically detect critical fall events and immediately alert medical professionals or caregivers. In recent years, various sensor devices have been used in FD systems, including smartphones~\cite{re5_Fall_detection_montoring}, accelerometers~\cite{Review1_Add2, re6_Near-Fall}, radio-frequency identification (RFID)~\cite{re10_A_Cloud-based}, cameras~\cite{re12_An_integrated}, and pressure sensors~\cite{re16_Vision-based}. Among these technologies, FD systems that use accelerometers have many advantages, such as high sampling rates, low cost, high efficiency, and portability~\cite{re17_The_Methods}. For example, S. Moulik~\textit{et al.}~\cite{Review1_Add2} proposed a fuzzy-based FD inference system to fuse the data of triaxial accelerometers from multiple sensors for accuracy improvement of FD.

Since a large amount of data from these sensors, machine learning (ML) and deep learning (DL) technologies have been widely applied to FD problems~\cite{Review1_Add1, Review1_Add3, Review1_Add4, re6_Near-Fall, re32_A_Novel, re33_Pre-Impact,re_add1_MLDL,re_add2_MLDL,re_add3_MLDL}. 
N. A. Choudhury~\textit{et al.}~\cite{Review1_Add1} reviewed and discussed the latest state-of-the-art human activity recognition (HAR) models with different applications, including ML, DL, fuzzy algorithms, etc.
FD is regarded as a classified problem, also known as HAR, which contains two possible outcomes: falls and activities of daily living (ADLs). 
, which can show similar peak data patterns. This similarity can result in false alarms (detecting a fall when the true activity is an ADL) or mis-detections (detecting an ADL when the true activity is a fall), leading to wasted medical resources and missed opportunities to treat injuries, respectively. Thus, the accuracy performance of FD is a critical issue. 
The main factors that impact the accuracy performance of the DL model include the type of used DL model, the quality of the dataset, data pre-processing, and the computational capability of the sensors or back-end computing etc.
In general, the computational capability of hardware is the key factor in training the DL model and supporting the computational cost for performance optimization~\cite{re24_A_Survey}.
A cloud center (CC) provides sufficient computational power and resources needed to apply DT technologies for performing FD with high accuracy~\cite{re10_A_Cloud-based,re22_A_smart,re24_A_Survey}.

Most of the HAR studies only focused on the accuracy measurements.
However, the main disadvantage of heavily relying on CC is the high transmission costs associated with high latency, network bandwidth, and privacy concerns~\cite{Review1_Add1,re24_A_Survey}.
N. A. Choudhury~\textit{et al.}~\cite{Review1_Add1} indicated real-time HAR system development is costly and lacks exploration because of the heavy bandwidth usage and real-time cloud or server architecture requirements.
Additionally, response latency in FD is a significant concern.
Therefore, N. A. Choudhury~\textit{et al.}~\cite{Review1_Add3} proposed an adaptive batch size-based model and combined hybrid convolutional neural networks (CNN) and long short-term memory (LSTM) as a CNN-LSTM framework for a smartphone sensor-based HAR system to improve recognition accuracy and computational efficiency.
Without data pre-processing and data augmentation, N. A. Choudhury~\textit{et al.}~\cite{Review1_Add4} also proposed an efficient and lightweight CNN-LSTM model for enhanced ADL classification on raw data of sensors in an uncontrolled environment.
However, these HAR works~\cite{Review1_Add3, Review1_Add4} only used the computational resource of edge devices (EDs) and neglected the latency between the ED and the mobile edge computing (MEC) (e.g., access point) or CC.
Furthermore, several works~\cite{re28_Artificiall,re29_Cloud-Fog, Review1_Add7, re_add5_EdgeCC, re_add4_EdgeCC} have focused on mobile edge computing (MEC) resources, such as access points, that communicate directly with edge devices (EDs) to reduce response latency compared to CC computing.
However, the computational capability of MEC is generally lower than that of the CC.
%A. Singh~\textit{et al.}~\cite{Review1_Add7} proposed an edge computing-based CNN model and compared the performances in terms of accuracy and latency with the cloud computing-based CNN model.
Therefore, the trade-off between accuracy and response latency needs to be considered to design the overall EDs-MECs-CC architecture is an essential issue.

\begin{figure}[t]
\centering
\includegraphics[width=0.45\textwidth]{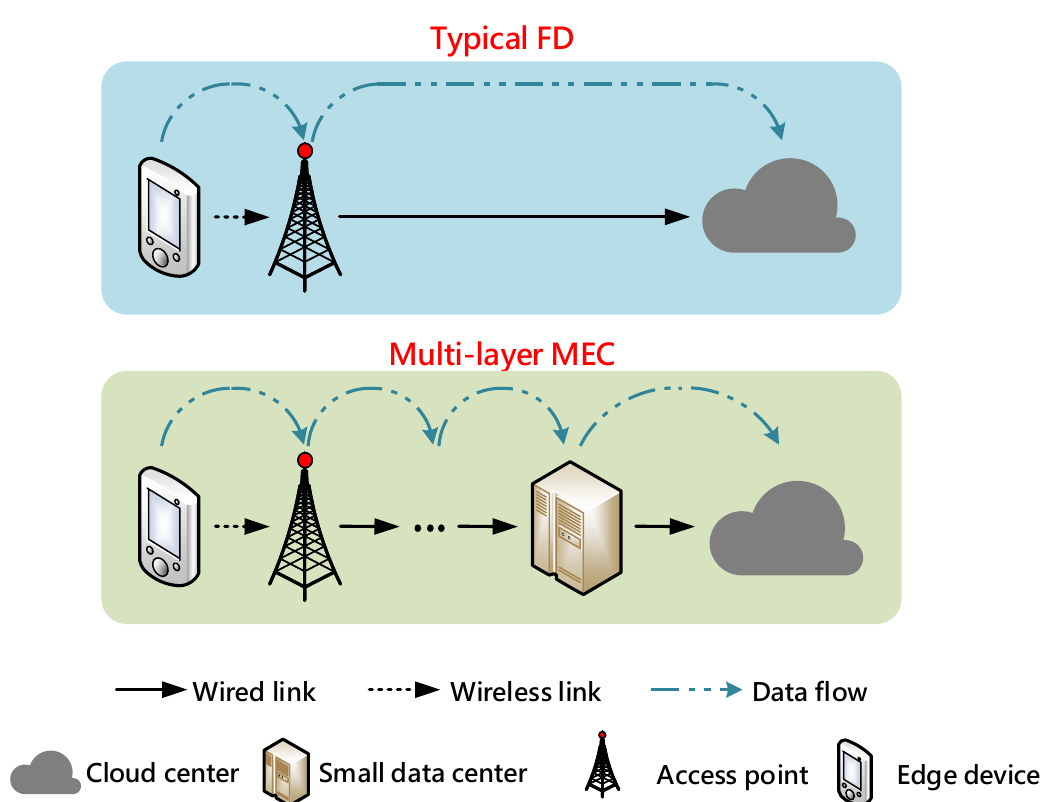}
\caption{Typical two-tier and the proposed MLMEC FD frameworks.}
\label{fig:Multi-layer MEC}
%\vspace{-0.1in}
\end{figure}

To address the issues above, we adopt the multilayer mobile edge computing (MLMEC) framework proposed in~\cite{re27_HetMEC} to balance the accuracy and latency of FD systems. Fig.~\ref{fig:Multi-layer MEC} illustrates the typical two-tier and the proposed MLMEC FD frameworks. In Fig.~\ref{fig:Multi-layer MEC}, a typical two-tier framework only relies on the computational capability of the ED or CC for FD and neglects the computational resources of MECs between ED and CC.
In contrast, the MLMEC framework is proposed to exploit the computing capabilities of the MECs in~\cite{re27_HetMEC, re30_HetMEC_6Gl}. Therefore, the proposed MLMEC framework for the FD system utilizes the computational resources of MECs between ED and CC and strikes a balance between detection performance and computational complexity.
The authors conducted a performance analysis of MLMEC systems in the 6G networks in~\cite{re30_HetMEC_6Gl} to enhance performance and efficiency, reduce task execution latency, and optimize resource utilization. Within the MLMEC framework, MECs (e.g., small data centers) typically exhibit varying computational capabilities. To bridge the performance gap between these capabilities, we employed knowledge distillation (KD)~\cite{re38_KD, re39_KD, re40_KDNN} to extract knowledge and enhance the accuracy of lightweight (student) models from relatively more powerful (teacher) pre-trained models.
KD is a common technique for model compression, aiming to reduce model complexity while maintaining accuracy~\cite{re41_Lightweight}.
KD compresses the knowledge of a larger and computationally expensive model (teacher) to
a smaller computationally efficient model (student). 
The idea of KD is to train the student model on a transfer set with soft targets provided by
the teacher model. 
The main contributions of this study are as follows:
\begin{itemize}
    \item We propose an MLMEC framework that manages the accuracy-latency trade-off using a threshold-based judgment for FD systems. 
    \item Using the KD approach, we bridge the performance gap between CC, MEC, and ED to significantly improve accuracy and reduce latency.
    \item We adopted the FLOPs computation to demonstrate the computational cost savings achieved by the proposed MLMEC framework compared to the typical two-tier cloud device architecture. 
    \item Two public FD datasets (FallAllD and Sisfall) were used to validate the effectiveness of the proposed model in achieving higher detection accuracy. 
    \item The source code (Python) is available on GitHub to train the three DL-based models (ResNet18, MobileNetV3, and CNN models) \footnote{The source codes are available at https://github.com/BoneZhou/MECKD}..  
\end{itemize}

The remainder of this paper is organized as follows.
In Sec.~\ref{sec:Related_Works}, the related works of a typical two-tier FD system are reviewed.
In Sec.~\ref{sec:Proposed_MLMEC}, the proposed MLMEC framework for FD is introduced. 
In Sec.~\ref{sec:DL-based FD}, the DL-based FD system with the KD approach is investigated. 
Sec.~\ref{sec:Performance metric} investigates the metrics used to evaluate the performance of the proposed MLMEC system.
Sec.~\ref{sec:Experimental results} presents the simulation results for dual- and triple-layer MLMEC systems with and without KD. Finally, Sec.~\ref{sec:Conclusion} summarizes the findings of this study.

\section{Related Works}
\label{sec:Related_Works}

 Several works~\cite{Review1_Add6, Review1_Add8, Review1_Add9, Review1_Add11} about deep learning-based fall detection with the KD approach are reviewed as follows. Most literature~\cite{Review1_Add6, Review1_Add8, Review1_Add9, Review1_Add11} only considers the single-stage KD approach.
T. X. Hoa~\textit{et al.}~\cite{Review1_Add6} proposed a novel 3DKD model combining KD and attention mechanisms to contribute more precise deployment models capable of operating efficiently on weak devices in computer vision tasks for FD.
J. Ni~\textit{et al.}~\cite{Review1_Add8} proposed a multi-modal approach in a vision sensor-to-wearable sensor with KD (VSKD) framework and designed a novel KD loss function to mitigate the modality variations between the vision and the sensor domain. It not only reduces the computational demands on wearable devices but also develops a DL-based model that closely matches the performance of the computationally expensive multi-modal approach.
H.-A. Rashid~\textit{et al.}~\cite{Review1_Add9} proposed a deep neural network (DNN) model with the KD and low bit-width quantization to fit models within lower memory hierarchy levels, reducing latency and enhancing energy efficiency on resource-constrained edge devices.
Z. Quan~\textit{et al.}~\cite{Review1_Add11} proposed a novel semantic-aware multimodal transformer fusion decoupled KD (SMTDKD) to guide video data recognition through the information interaction among different wearable sensor data and visual sensor data.

The most related state-of-the-art (SOTA) work is our previous study~\cite{add7_Liu}, which proposed a novel pre-impact fall detection via CNN-ViT knowledge distillation, namely PreFallKD, to strike a balance between detection performance and computational complexity. The CNN-ViT trains a CNN student model using a vision transformer (ViT)~\cite{Vision_Transformer} teacher model to enhance the performance of the CNN student model through the KD approach. The experiment results in~\cite{add7_Liu} show that the latency of CNN is much less than other SOTA models, such as CNN-LSTM and ViT models, etc. However, the training cost of the ViT model requires three types of sensor data, including triaxial acceleration data, triaxial gyroscope data, and triaxial Euler angle data from the KFall dataset~\cite{KFall} for convergence. In contrast, the training costs of ResNet, MobileNet, and CNN models in the proposed MLMEC require only a single type of sensor data from FallAllD~\cite{re25_FallAllD} and SisFall~\cite{re26_SisFall} datasets. Therefore, it is difficult to compare the proposed MLMEC with KD and PreFallKD~\cite{add7_Liu} under unequal conditions. 

However, when there is a significant gap in performance between the teacher and student models, the effectiveness of the single-stage KD is limited~\cite{re39_KD, Review1_Add10}. A. H. Saleknia~\textit{et al.}~\cite{Review1_Add10} proposed an innovative multi-stage KD framework that takes advantage of a mid-size teacher assistant (TA) network to narrow the computational gap between student and teacher networks for computer vision task. Especially, the computational capabilities of edge servers are inconsistent in general heterogeneous networks~\cite{re27_HetMEC, re30_HetMEC_6Gl}. Thus, the gap in performance between the teacher and the student models is difficult to anticipate accurately. For more robust KD effectiveness, the multi-stage KD approach is proposed to bridge the performance gap between teacher and student models by the TA models~\cite{re39_KD}.

To the best of our knowledge, this work is the first to integrate the multilayer MEC framework and the multi-stage KD approach~\cite{re39_KD} for FD in wearable devices.
The trade-off between accuracy and latency is considered an essential evaluation metric for FD systems. Accuracy improvement reduces the likelihood of false alarms and missed detections, and increases medical resource utilization and treatment opportunities. 
However, even with high accuracy, reducing response latency remains a crucial issue.
The proposed MLMEC with KD is to enhance both metrics simultaneously.
Through KD training from the upper-layer MEC or cloud center, the accuracy of the lower-layer MEC is improved in the MLMEC framework.
When the accuracy of the lower layer is increased, the computation of the upper layers is reduced, resulting in reduced response latency.
Therefore, the design of hyperparameters for the neural models at different layers of the MEC is crucial.
 %Moreover, the proposed MLMEC framework can not only be applied to fall detection but also to other bio-signal processing, such as cardiac arrhythmias detection using electrocardiography (ECG) signals~\cite{re47_SRECG}.

\section{Proposed MLMEC Framework for FD}
\label{sec:Proposed_MLMEC}

The proposed MLMEC framework with the KD approach for FD includes the bottom-layer EDs, the middle-layer MEC servers, and the top-layer CC server. All the components in the MLMEC are equipped with different sizes of DL models. Based on the computational capability of the component, each layer is assigned a DL model with specific accuracy. In general, the top-layer CC server is equipped with the most computationally expensive DL model, the lowest-layer MEC is equipped with the lightest DL model, and the middle-layer MECs are equipped with the middle-sized DL model. The KD training process is adopted in the MLMEC framework to enhance the accuracy performance of the lighter-sized DL models (student) by the relatively more powerful pre-trained DL models (teacher). The author in~\cite{re40_KDNN} demonstrated that the KD training process is highly effective in transferring knowledge from either an ensemble or a large, extensively regularized model to a smaller, distilled model. It also showed substantial improvements obtained through distilling a student model of equivalent size, which is much easier for deployment in the MLMEC framework. However, the effectiveness of knowledge distillation might be limited when there is a significant gap in performance between the teacher and student models. Therefore, teacher-assistant KD (TAKD) was proposed in~\cite{re39_KD}, which involves incorporating intermediate models as teacher assistants (TAs) between the teacher and the student to bridge the performance gap. The authors in~\cite{re39_KD} conducted a theoretical analysis demonstrating that the TAKD (teacher-TA-student model) outperforms the baseline KD (teacher-student model) and without KD (original student model).

\begin{figure*}[h]
\centering
      \includegraphics[width=1\textwidth]{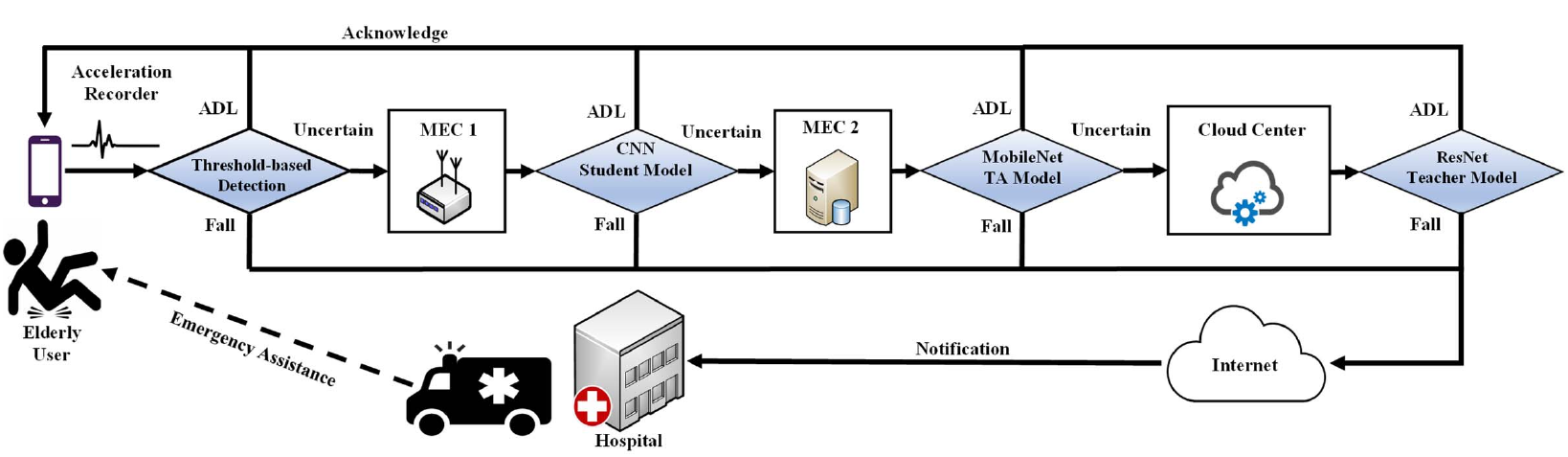}
   \caption{The application scenario of the proposed MLMEC FD system.}
\label{fig:MECKD_system_F1}
\end{figure*}

Fig.~\ref{fig:MECKD_system_F1} illustrates the operational flow and application scenario of the proposed FD system in the MLMEC system with KD. Our goal is to enhance the accuracy of the FD system and reduce the latency. The users used a smartphone equipped with a threshold-based FD~\cite{re18_The_Methods, re19_An_analysis}. 
The threshold-based detection has low complexity and is easy to implement in wearable sensors. 
The details of the threshold-based FD are introduced in Sec.~\ref{subsec:Feature extraction with threshold-based classification}. The smartphone reads acceleration values from the accelerometer and assesses whether the detected value corresponds to a fall, ADLs, or remains uncertain. In the case of a fall detection, an alert is sent to the hospital for emergency ambulance assistance via the Internet. For ADL detection, the system records only the intense signal in the acceleration recorder. In case of an uncertain detection, the signal is forwarded to MEC 1 for a more accurate determination. If the detection by MEC 1 remains uncertain, the signal is transmitted by MEC 2. If the detection of MEC 2 is still uncertain, the signal is routed to the upper-layer MECs for a more precise determination than that provided by the lower-layer MECs. If all the detection results of MECs are uncertain, the signal is sent to the CC for a final decision. The proposed FD system architecture is deployed in the MLMEC with KD. In Fig.~\ref{fig:MECKD_system_F1}, CC, MEC 1, and MEC 2 are equipped with DL models, specifically ResNet~\cite{re46_ResNet} (large), MobileNetV3~\cite{re45_MobileNetV3} (medium), and basic convolutional neural networks (CNNs)~\cite{re_add6_CNN} (small), respectively. 

By adopting the KD technique, the accuracy of the TA model in MEC 2 is enhanced by using the teacher model in CC, and the accuracy of the student model in MEC 1 is improved by the TA model in MEC 2. In addition, the TA models can be deployed to the different MEC layers between the teacher model and the student model. This effectively bridges the performance gap among CC, MECs, and EDs. By TA-based KD, the amount of data labels is reduced, enhancing the usability of the FD model in real-world applications.
In particular, the MLMEC framework is suitable for deploying TA-based KD~\cite{re39_KD}.
Compared to the traditional two-tier cloud-device architecture, the proposed MLMEC framework saves significant computational costs in our experiments.

\section{DL-based FD with KD in MLMEC Framework}
\label{sec:DL-based FD}
Our goal is to propose an MLMEC framework that strikes a good trade-off between the accuracy and latency of the FD system. Based on the MLMEC framework, the TAKD is introduced to improve the accuracy performance of the lower MEC layer from the upper MEC  layer, while the latency of the FD system is reduced.

\begin{figure}[htbp]
\centering
\includegraphics[width=0.48\textwidth]{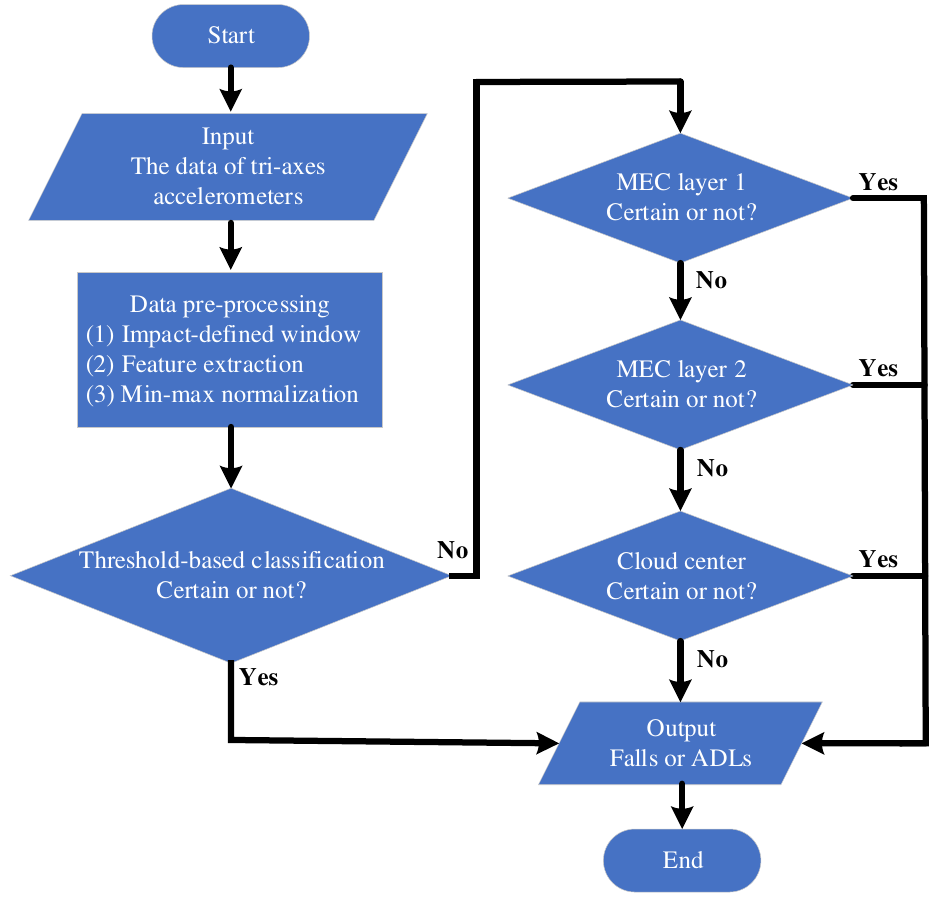}
\caption{The flow chart of the proposed MLMEC FD system.}
\label{fig:Training process}
\end{figure}

The proposed MLMEC framework for the FD system is illustrated in Fig.~\ref{fig:Training process}. The approach consists of four stages: 1) data pre-processing; 2) threshold-based classification; 3) DL-based FD in the proposed dual-layer and triple-layer MEC system with the KD approach; and 4) threshold-based upward judgment. In the data pre-processing stage, the data is segmented using an impact-defined window with a fixed window size. The threshold-based classification is performed at the ED based on the maximum and minimum spatial momentum values (threshold) between ADLs and fall states. The advantage of this approach is that it provides real-time feedback and subsequent processing for clear fall situations.
The details of threshold-based classification of ED for feature extraction will be presented in Sec.~\ref{subsec:Feature extraction with threshold-based classification}.
The uncertain detection results of the previous MEC layers are processed using DL models in the subsequent MECs and CC for detection. The loss function design of the DL model in the proposed dual-layer and triple-layer MEC system with the KD approach will be presented in Sec.~\ref{sec:Loss Function of Dual-layer and Triple-layer MECs}. As for the decision to transfer upward to the next MEC layer, we propose the threshold-based judgment for MECs in Sec.~\ref{sec:Threshold-based Upward judgment}.

\begin{figure}[t]
\centering
\includegraphics[width=0.5\textwidth]{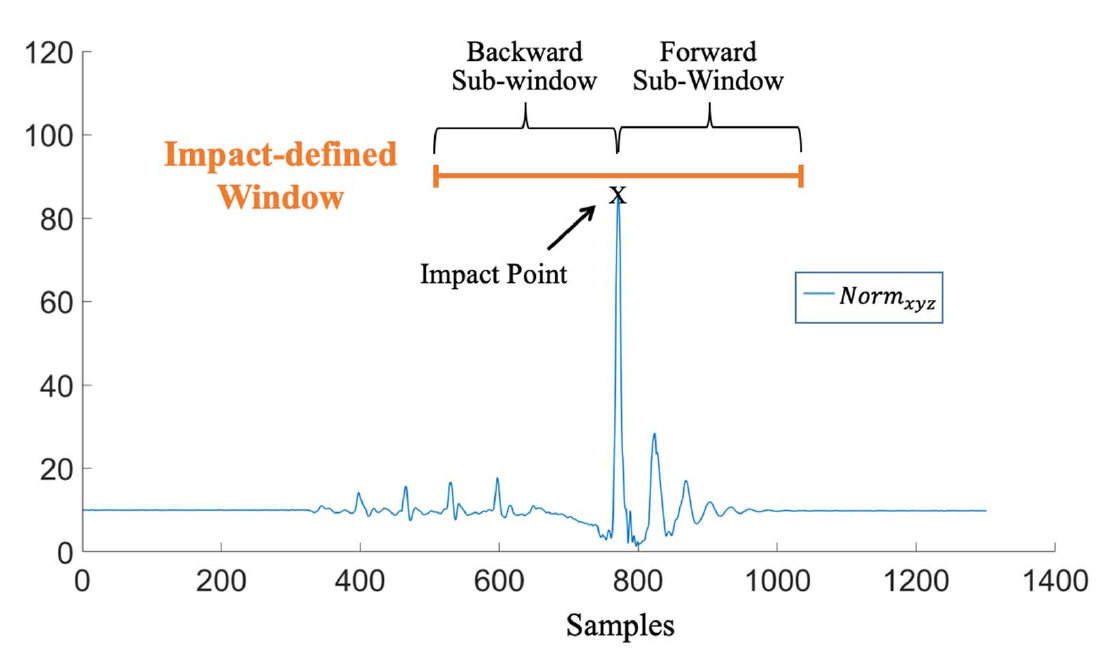}
\caption{Impact-defined window on a fall trial~\cite{re19_An_analysis, Review1_Add15}, where the X-axis represents the sampling time (unit: second), and the Y-axis represents the acceleration momentum (unit: gram).}
\label{fig:Impact_defined}
\end{figure}

The impact-defined window is a common method for pattern detection. The window size is determined by the timing of the impact point to capture falling patterns (before, during, and after the fall).
Fig.~\ref{fig:Impact_defined} illustrates the impact-defined window for a fall signal.
A peak signal is generated at the time of impact, and time frames before and after the peak point are searched to identify the sub-windows before and after the fall, which are used to form a fixed-size window for detecting the fall patterns.

\subsection{Data Pre-Processing}
\label{subsec:Pre-processing}
\subsubsection{Impact-Defined Window}
\label{sususec:Impact-defined window}

Assuming that the signal from the accelerometer is defined as $S=\{s_{j}|j=1,2\cdots,n_{s}\}$, where $n_{s}$ is the total number of samples.
Each sample point includes XYZ 3-axis acceleration $s_{j}=\{a_{x_{j}},a_{y_{j}},a_{z_{j}}\}$.
Firstly, the sample with the maximum $Norm_{xyz}$ value is identified as the impact point $s_{I}$, and the $Norm_{xyz}$ formula for the $j$-th sample $s_{j}$, which is shown as
\begin{equation}
        Norm_{xyz}(s_{j})=\sqrt{a_{x_{j}}^{2}+a_{y_{j}}^{2}+a_{z_{j}}^{2}}.
        \label{eq3:Norm}
\end{equation}
The maximum boundary value for identifying falls and the minimum boundary value for identifying non-falls are extracted based on the $Norm_{xyz}$ distributions of falls and ADLs. 
Based on the impact point $s_{I}$, the sub-windows before and after the impact point are defined as $W_{f}=\{S_{I+1,}\cdots, S_{I+WS_{f}-1}, S_{I+WS_{f}}\}$ and $W_{b}=\{S_{I-WS_{b}}, S_{I+WS_{b}+1},\cdots, S_{I-1}\}$.
The $WS_{f}$ and $WS_{b}$ are described the sub-window sizes of $W_{f}$ and $W_{b}$. The details of window determination were introduced in~\cite{re18_The_Methods, re19_An_analysis, re49_An_Analysis}. 

\subsubsection{Feature Extraction}
\label{sususec:Feature extraction}

For robust fall detection, there are two critical issues that need to be tackled as follows:
\begin{enumerate}
\item \emph{Variability}: Falls may happen suddenly in various directions and forms in activity transitions of daily living~\cite{Review1_Add16}.
\item \emph{Ambiguity}: Since some features of ADLs are similar to the features of falls, which may confuse the FD system, such as the person jumping or running in daily living leads to strong impact and energy, which may be misidentified as falls, and vice versa~\cite{re17_The_Methods}.
\end{enumerate}

Based on the previous work~\cite{re17_The_Methods}, we adopt the nine commonly used time-domain statistical features as the extracted features from the impact-defined window, including mean, standard deviation, variance, maximum, minimum, range, kurtosis, skewness, and correlation coefficient~\cite{Review1_Add17}. These nine features are extracted from the Euclidean norm of tri-axial acceleration, the Euclidean norm of acceleration on the coronal plane, and the Euclidean norm of acceleration on a horizontal plane. Tab.~\ref{tb:extracted features} shows the 54 features used in this study. These nine features are widely used in FD~\cite{re5_Fall_detection_montoring} and HAR~\cite{Review1_Add18}.

\begin{table}[t]
\centering
\caption{List of the extracted features\cite{re19_An_analysis, Review1_Add15}.}
\scalebox{0.9}{
      \begin{tabular}{|c|l|}
      \hline
       \textbf{No.} & \textbf{Time-domain statistical features}\\\hline
        $f_{1}$-$f_{6}$ & Mean of $a_{x}$, $a_{y}$, $a_{z}$, $a_{norm}$, $a_{verti}$, $a_{hori}$\\\hline
        $f_{7}$-$f_{12}$ & Standard deviation of $a_{x}$, $a_{y}$, $a_{z}$, $a_{norm}$, $a_{verti}$, $a_{hori}$\\\hline
        $f_{13}$-$f_{18}$ & Variance of $a_{x}$, $a_{y}$, $a_{z}$, $a_{norm}$, $a_{verti}$, $a_{hori}$\\\hline
        $f_{19}$-$f_{24}$ & Maximum of $a_{x}$, $a_{y}$, $a_{z}$, $a_{norm}$, $a_{verti}$, $a_{hori}$\\\hline
        $f_{25}$-$f_{30}$ & Minimum of $a_{x}$, $a_{y}$, $a_{z}$, $a_{norm}$, $a_{verti}$, $a_{hori}$\\\hline
        $f_{31}$-$f_{36}$ & Range of $a_{x}$, $a_{y}$, $a_{z}$, $a_{norm}$, $a_{verti}$, $a_{hori}$\\\hline
        $f_{37}$-$f_{42}$ & Kurtosis of $a_{x}$, $a_{y}$, $a_{z}$, $a_{norm}$, $a_{verti}$, $a_{hori}$\\\hline
        $f_{43}$-$f_{48}$ & Skewness of $a_{x}$, $a_{y}$, $a_{z}$, $a_{norm}$, $a_{verti}$, $a_{hori}$\\\hline
        $f_{49}$-$f_{51}$ & Correlation coefficient between each pair of $a_{x}$, $a_{y}$, $a_{z}$\\\hline
        $f_{52}$-$f_{54}$ & Correlation coefficient between each pair of $a_{norm}$, $a_{verti}$, $a_{hori}$\\\hline
        \end{tabular}}
     \label{tb:extracted features}
\end{table} 

\subsubsection{Min-Max Normalization}
\label{sususec:Min-Max Normalization}
Once the impact-defined window is determined, the data is processed using min-max normalization to reduce scaling effects during the training phase. By applying min-max normalization, each element of the data sequence $S=\{s_{i}|i=1,2\cdots, I\}$ is normalized to a range between 0 and 1, which is shown as
\begin{equation}
       S_{i}^{nom}=\frac{s_{i}-s_{min}}{s_{max}-s_{min}}.
        \label{eq4:Sj}
\end{equation}
where $s_{max}$ and $s_{min}$ are the maximum and minimum values of $S$, respectively.
In addition, we have compared min-max normalization and z-score standardization (standard scalar standardization) in our experiments for fall detection.

\subsection{Threshold-Based Classification}
\label{subsec:Feature extraction with threshold-based classification}
\begin{figure}[t]
\centering
    \subfloat[\label{subfig:ADLa}]{%
      \includegraphics[width=0.49\textwidth]{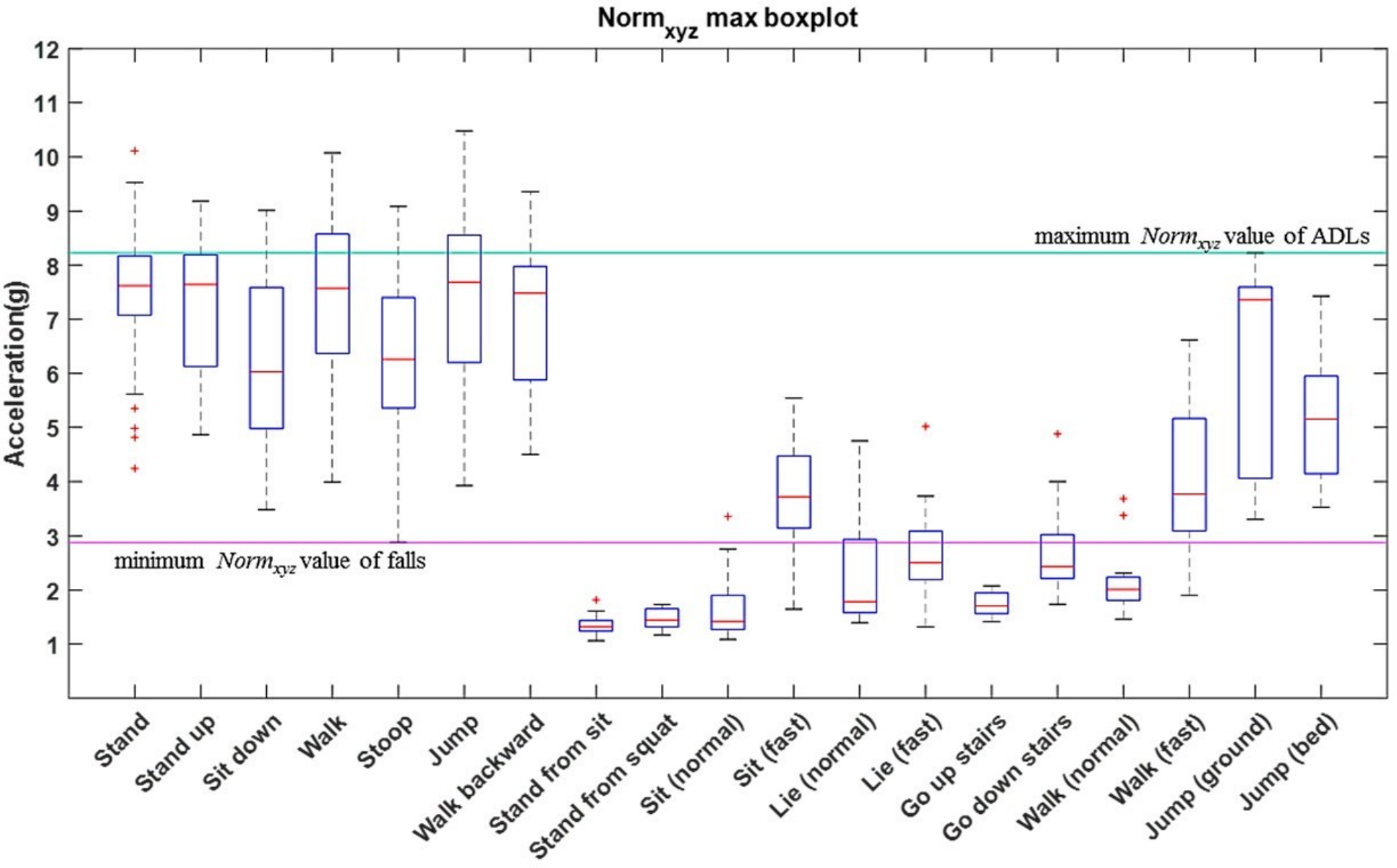}
      }
      \hfill
    \subfloat[\label{subfig:ADLb}]{%
      \includegraphics[width=0.49\textwidth]{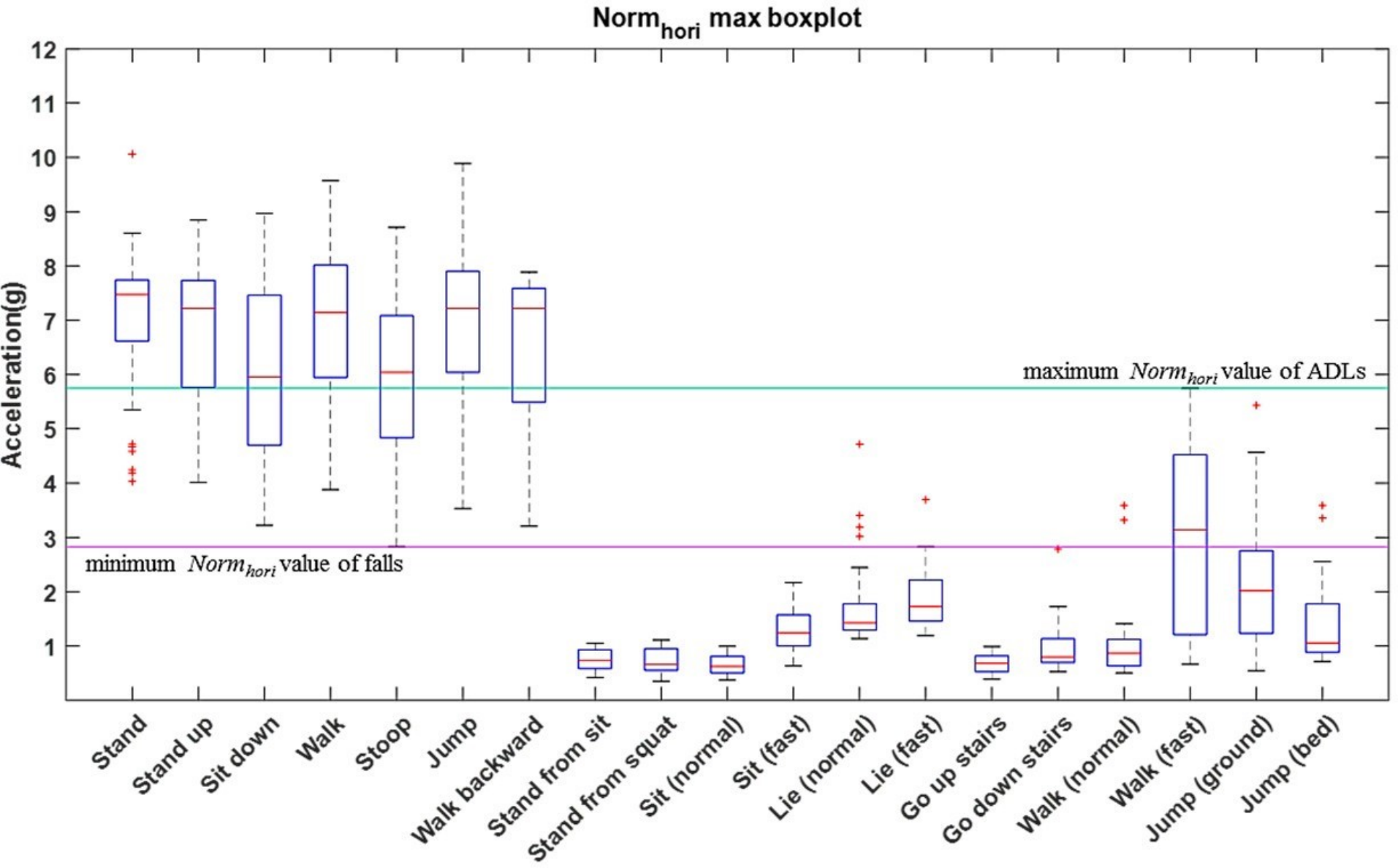}
      }
   \caption{The illustration of the $Norm_{xyz}$ and $Norm_{hori}$ distribution of falls and ADLs~\cite{re18_The_Methods, re19_An_analysis} (a) The boxplots of maximum and minimum values of $Norm_{xyz}$ for falls and ADLs; (b) The boxplots of maximum and minimum values of $Norm_{hori}$.}
\label{fig:Norm}
\end{figure}

In our study, we categorized all human actions into ADLs and falls. The features of falls and ADLs are extracted by the threshold-based classification. An initial threshold-based detection using maximum and minimum spatial momentum values is employed. The maximum and minimum spatial momentum values are determined by two types of data features, $Norm_{xyz}$ and $Norm_{hori}$, that are extracted from each data frame. $Norm_{xyz}$ is calculated by Eq.~\eqref{eq3:Norm} and characterizes the spatial variation of acceleration during the falling interval. 
$Norm_{hori}$ is defined as the Euclidean norm of acceleration for capturing the change of velocity in the horizontal plane of the body and is calculated as follows:
\begin{equation}
       Norm_{hori}(S_{j})=\sqrt{a_{y_{j}}^{2}+a_{z_{j}}^{2}}.
        \label{eq5:Norm_hori}
\end{equation}
The threshold is determined by the maximum and minimum values of $Norm_{xyz}$ and $Norm_{hori}$~\cite{re18_The_Methods, re19_An_analysis} as shown in Fig.~\ref{fig:Norm}. 
Fig.~\ref{fig:Norm} shows the $Norm_{xyz}$ and $Norm_{hori}$ distributions of falls and ADLs, where the left seven activities are falls and the right twelve activities are ADLs, the green line is determined by the maximum value of $Norm_{xyz}$ ($Norm_{hori}$) of ADLs, and the purple line is determined by the minimum value of $Norm_{xyz}$ ($Norm_{hori}$) of falls. 

Assume that there is a training set $F_{train}=\{(f_{i}^{train})|i=1,2\cdots, N_{train}\} $, where $N_{train}$ is the total number data frames, and there is a testing set $F_{test}=\{(f_{i}^{test})|i=1,2,\cdots, N_{test}\}$, where $N_{test}$ is the total number of the set $F_{test}$, two sets of maximum $Norm_{xyz}$ and $Norm_{hori}$ corresponding to $F_{train}$ are presented as $V_{train}=\{(v_{i}^{train}|i=1,2,\cdots, N_{train})\}$ and $w_{train}=\{(w_{i}^{train}|i=1,2,\cdots, N_{train})\}$, respectively. Based on the distribution of $V_{train}$ and $W_{train}$, two thresholds, $T_{fall}^{Norm_{xyz}}$ and $T_{fall}^{Norm_{hori}}$, for absolute falls are determined, where $T_{fall}^{Norm_{xyz}}$ and $T_{fall}^{Norm_{hori}}$ are the maximum $Norm_{xyz}$ and $Norm_{hori}$ values of ADLs, respectively. Additionally, two thresholds, $T_{ADL}^{Norm_{xyz}}$ and $T_{ADL}^{Norm_{hori}}$, for the absolute ADL identification are determined, where$T_{ADL}^{Norm_{xyz}}$ and $T_{ADL}^{Norm_{hori}}$ are the minimum $Norm_{xyz}$ and $Norm_{hori}$ values of falls, respectively. Similarly, the maximum and minimum values of $Norm_{xyz}$ and $Norm_{hori}$ corresponding to $F_{test}$ are presented as $V_{test}=\{(v_{i}^{test})|i=1,2,\cdots,N_{test}\}$ and $W_{test}=\{(w_{i}^{test})|i=1,2,\cdots,N_{test}\}$, respectively. Finally, the threshold-based classification (TC) applied to the test set is defined as follows:
\begin{equation}
\begin{aligned}
        &TC(v_{i}^{test},w_{i}^{test})=\\
        &\begin{cases}
        {\rm Fall, if }\ v_{i}^{test} > T_{fall}^{Norm_{xyz}} {\rm and}\ w_{i}^{test} > T_{fall}^{Norm_{hori}} \\
        {\rm ADL, if }\ v_{i}^{test} < T_{fall}^{Norm_{xyz}} {\rm and}\ w_{i}^{test} < T_{fall}^{Norm_{hori}} \\
        {\rm Uncertain,\ others.}
        \end{cases}
        \label{eq6:TC}
\end{aligned}
\end{equation}
Therefore, three regions are distinguished: the absolute boundary of falls and ADLs, as well as the range of uncertain data that will be transferred to the MEC.

\begin{figure*}[tb]
\centering
    \subfloat[\label{subfig:Dual-layer}]{%
      \includegraphics[width=0.7\textwidth]{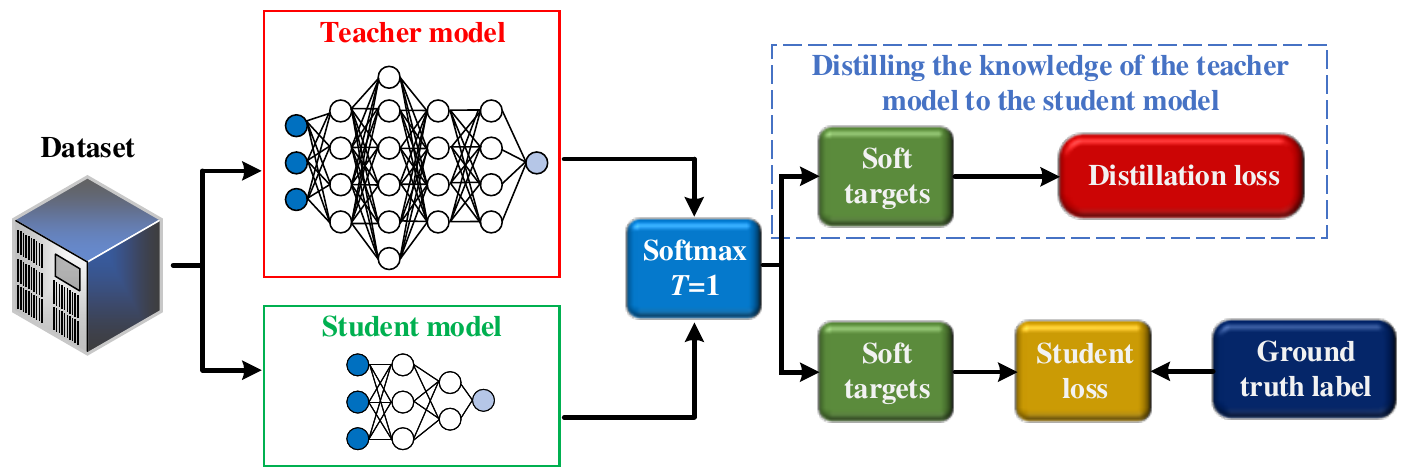}
      }
      \hfill
    \subfloat[\label{subfig:Triple- layer}]{%
      \includegraphics[width=0.7\textwidth]{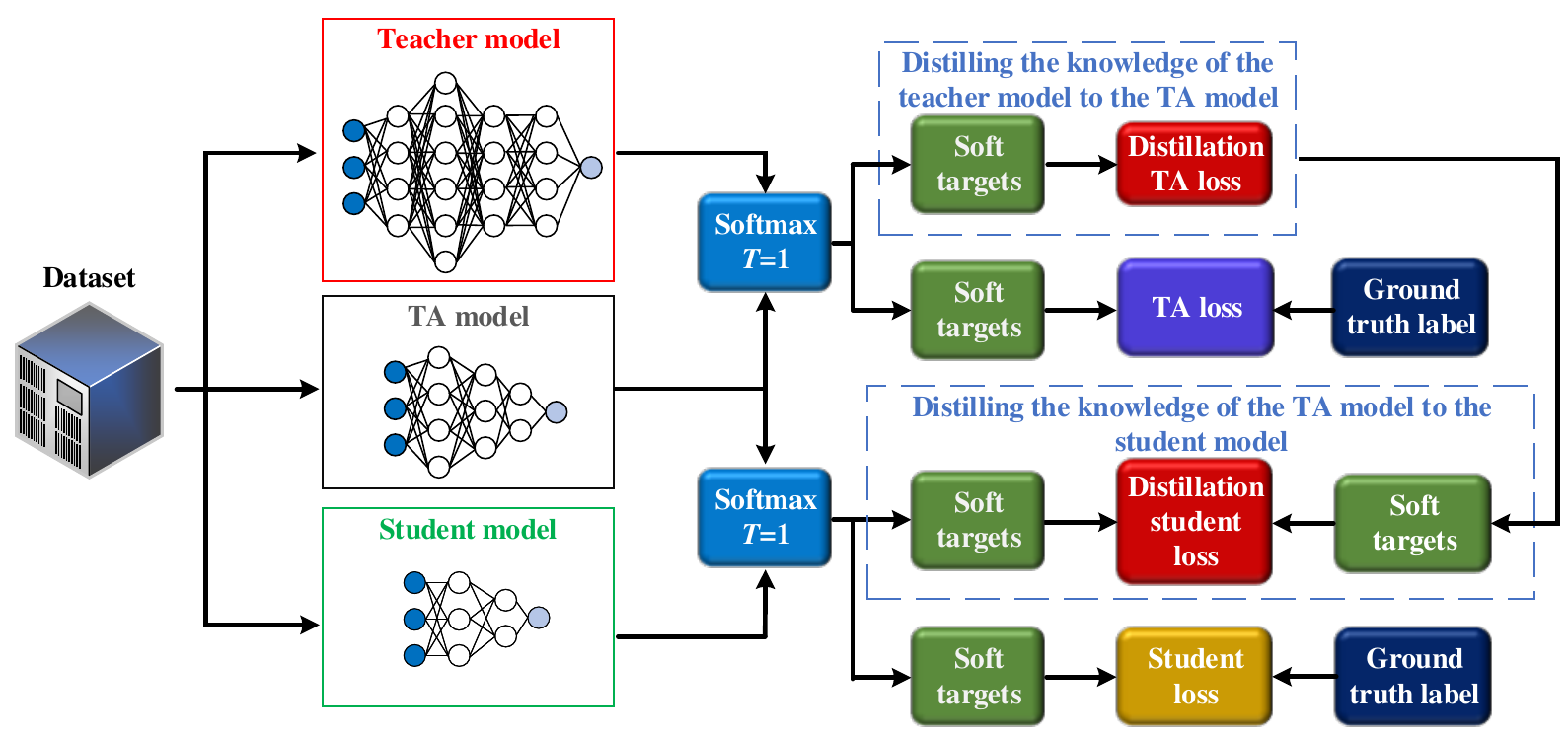}
      }
   \caption{The multilayer MECs with the KD approach. (a) Dual-layer (b) Triple-layer.}
\label{fig:KD model architecture}
  \end{figure*}

\subsection{Loss Functions of Dual-layer and Triple-layer MECs}
\label{sec:Loss Function of Dual-layer and Triple-layer MECs}

In this subsection, the loss functions of dual-layer and triple-layer MECs are analyzed. 
The architectures of dual-layer and triple-layer MECs with the KD approach are shown in Fig.~\ref{fig:KD model architecture}.

As shown in~\ref{fig:KD model architecture}(a), the teacher model is pre-trained based on the dual-layer MEC architecture.
Then, we use the pre-trained teacher model and ground truth (label) to train the student model. 
By this KD approach, the weights of the teacher models are distilled into the student model.
This KD approach improves the performance of the student model.

In general, supervised learning uses cross-entropy loss to measure the mismatch between the output of the student model and the ground truth (label) as follows:
\begin{equation}
       loss = {\cal H} (g, Student).
        \label{eq7:Loss}
\end{equation}
where $g$ represents the ground-truth value, $Student$ represents the output of the student model, and ${\cal H}(p, q) = -\mathbb{E}_{p}(\log (q))$. 

In the dual-layer MECs architecture (teacher-student model), we use the Kullback–Leibler (KL) divergence loss to match the softened outputs of the teacher and student models as follows
\begin{equation}
\begin{aligned}
       KL&\left(\frac{Teacher}{T},\frac{Student}{T}
       \right)=
       \\&\left(\frac{Student}{T}\right)\left[\log\left(\frac{Student}{T}\right)-\log\left(\frac{Teacher}{T}\right)\right].
        \label{eq8:KL}
\end{aligned}
\end{equation}
where $Teacher$ represents the predicted values of the teacher model, $Student$ represents the predicted values of the student model, KL represents the KL-divergence between the output probabilities of the $Teacher$ and $Student$, and $T$ is referred to as an additional control parameter between the ground truth (label) and the soft targets from the output of the teacher model.

The loss function of the student model is as follows
\begin{equation}
\begin{aligned}
       loss_{dual}=&\lambda T^{2}\times KL\left(\frac{Teacher}{T},\frac{Student}{T}\right)
       \\&+(1-\lambda)loss),
        \label{eq9:KDLoss_dual}
\end{aligned}
\end{equation}
where $\lambda$ is a hyperparameter that adjusts the trade-off between two loss functions. 

According to~\cite{re39_KD}, the TAKD is used to improve the performance of the student model of the teacher-TA-student model, when the performance gap between the teacher and student is large. As shown in~\ref{fig:KD model architecture}(b), the weights of the pre-trained teacher model are extracted and distilled into the TA model. Then, the weights of the pre-trained TA model are further distilled into the student model. 

In the triple-layer MECs architecture, the KD loss function in the teacher-TA-student model is formulated. Firstly, we formulate the first KL divergence between the teacher model and the TA model. Subsequently, the second KL divergence is calculated for the distilled TA model and the student model. The first KL divergence is substituted into the second KL divergence, as follows:
\begin{equation}
\begin{aligned}
       KL&\left(\frac{Teacher}{T},\frac{TA}{T},\frac{Student}{T}\right)=
       \left(\frac{Student}{T}\right)
       \\&\times\left \{ \log\left(\frac{Student}{T}\right)-\log\left(\frac{TA}{T}\right) \right.
       \left.
       \left[\log \left(\frac{TA}{T}\right)-\log \left(\frac{Teacher}{T}\right)\right] \right \},
        \label{eq10:KL_Tri}
\end{aligned}
\end{equation}
where $TA$ represents the predicted values of the TA model. Then, we substitute the obtained values into Eq.~\eqref{eq11:KDLoss_Tri} to calculate the loss function of the student model in the triple-layer MECs architecture as follows
\begin{equation}
\begin{aligned}
       loss_{tri}=&\lambda T^{2}\times KL\left(\frac{Teacher}{T},\frac{TA}{T},\frac{Student}{T}\right)
       \\&+(1-\lambda)loss.
        \label{eq11:KDLoss_Tri}
\end{aligned}
\end{equation}

\subsection{Threshold-Based Judgment for Uncertain Data}
\label{sec:Threshold-based Upward judgment}

For processing the uncertain data of the previous layer MEC, a threshold-based judgment is proposed to distinguish absolute falls, ADL, and uncertain data. We define two threshold values, namely maximum and minimum upward thresholds $T_{fall}^{Max}$ and $T_{fall}^{Min}$, which are preset to define the range of the absolute falls, ADL, and uncertain data. 

If the result of the judgment remains uncertain, the data is sent to the larger model in the upper-layer MEC for a more precise detection. After the softmax function in the DL model of the $n$-th layer MEC, the softened output is obtained as follows:
\begin{equation}
        v_{n}^{test} = {\rm softmax}\left(\frac{Layer_{n}}{T}\right), 
        \label{eq1:softmax}
\end{equation}
where $Layer_{n}$ represents the predicted result of the $n$-th layer MEC. Then, we input the softened output of the $n$-th layer MEC to the threshold-based upward judgment ($TQ$), which is defined as follows
\begin{equation}
        TQ(v_{j}^{test})=\begin{cases}
        {\rm Fall, \; if }\ v_{n}^{test} > T_{fall}^{Max}\\
        {\rm ADL, \; if }\ v_{n}^{test} < T_{fall}^{Min}\\
        {\rm Uncertain, \; if }\  T_{fall}^{Max} > v_{n}^{test} > T_{fall}^{Min}.
        \end{cases}
        \label{eq2:TQ}
\end{equation}

\section{Performance Evaluation}
\label{sec:Performance metric}
The performance evaluation of the FD system was carried out using the leave-one-out cross-validation method. In this approach, data from one subject served as the test set while the data from the other subjects were used as the training set. This process was repeated $k$ times, where $k$ is the total number of subjects, ensuring that each subject’s data was used as the test set. After all iterations, the overall test results are aggregated, and the average performance of the $k$ Folds is considered as the final output for evaluating the performance of the FD system.

Applying the concept of a confusion matrix, four performance metrics are introduced, including accuracy ($ACC$), precision ($PRE$), recall ($REC$), and F1-score ($F1$), to measure the detection performance. The definitions of performance metrics are as follows:
\begin{align}
       ACC &= \frac{TP+TN}{TP+FP+TN+FN},\\
       PRE &= \frac{TP}{TP+FP},\\
       REC &= \frac{TP}{TP+FN},\\
       F1 &= \frac{PRE\times REC}{PRE+REC},
\end{align}
where True Positive ($TP$) refers to instances where fall signals are correctly identified as falls. True Negative ($TN$) indicates instances where activities of daily living (ADLs) signals are correctly identified as ADLs. False Positive ($FP$) represents instances where ADL signals are incorrectly labeled as falls, while False Negative ($FN$) denotes instances where fall signals are incorrectly labeled as ADLs.

In~\cite{re27_HetMEC}, the latency evaluation of each MEC of the MLMEC framework is calculated. The latency of the MLMEC architecture is defined as the total computation and transmission time from the ED to the CC. From ED to CC, each layer of the MLMEC framework processes incoming data from the previous layer. Therefore, the latency between ED and CC is expressed as
\begin{equation}
       L_{n}=\sum_{j=1}^{M_{n-1}}\sum_{i\in Q_{n-1}^{j}} \left [\frac{s_{n}^{i}b_{n}^{i}}{\theta_{n}^{i}}+\frac{\rho s_{n}^{i}\lambda_{n}^{i}+(1-s_{n}^{i})\lambda_{n}^{i}+\beta_{n}^{i}}{\varphi_{n}^{j,i}}\right].
        \label{eq25:Ln}
\end{equation}
The notation definitions are provided in Table~\ref{tb_Summary of key notation}.
\begin{table}[htbp]
\centering
\caption{Notations of latency calculation.}
\scalebox{0.75}{
      \begin{tabular}{|c|l|}
      \hline
        \textbf{Variable} & \textbf{Definition} \\
        \hline
        \textbf{$M_{n-1}$} & \textbf{The number of devices on each layer $n-1$} \\
        \hline
        \textbf{$Q_{n-1}^{j}$} & \textbf{The number of child nodes connected with the parent node $j$ on layer $n-1$} \\\hline
        \textbf{$\lambda_{n}^{i}$} & \textbf{Data generation speed of ED $i$} \\
        \hline
        \textbf{$\beta_{n}^{i}$} & \textbf{The total volume of the processed data received by node $i$ on layer $n$} \\
        \hline
        \textbf{$s_{n}^{i}$} & \textbf{The (equivalent) task division percentage at node $i$ on layer $n$} \\
        \hline
        \textbf{$\theta_{n}^{i}$} & \textbf{The computing capacity of node $i$ on layer $n$} \\
        \hline
        \textbf{$\varphi_{n}^{j,i}$} & \textbf{The transmitting capacity of node $i$ on layer $n$ to its parent node $j$} \\
        \hline       
    \end{tabular}}
     \label{tb_Summary of key notation}
\end{table}

\section{Experimental Results and Discussions}
\label{sec:Experimental results}

%\subsection{Open datasets}
%\label{subsec:Open Datasets}

In our simulations, FallAllD~\cite{re25_FallAllD} and SisFall~\cite{re26_SisFall} datasets were chosen to validate the proposed MLMEC FD system and the KD approach for managing the trade-off between accuracy and latency. 
These two datasets were collected from real-life fall types and ADLs, and the sampling rates are 200 Hz.
We set the window sizes of the sub-windows before and after the impact point (i.e., $WS_{f}$ and $WS_{b}$), respectively, to be 2 seconds (s) and 1.23 s for the FallAllD dataset, and 2 s and 1.44 s for the SisFall dataset. Our previous study~\cite{re19_An_analysis} has shown that such window size settings achieved optimal performance of the proposed FD systems for these two public datasets.
Pytorch 1.13.0 is used on a 64-bit Windows 10 PC with an Intel 13th Gen I7-13700K CPU 3.40GHz and an Nvidia RTX4080 GPU with 16GB dedicated memory. The data pre-processing and DL-based modeling were performed in an Anaconda environment.

\subsection{Open Datasets}
\label{subsec:Open Datasets}
%The public datasets for wearable FD systems include FallAllD~\cite{re25_FallAllD}, SisFall~\cite{re26_SisFall}, UMAFall~\cite{re23_UMAFall}, and UPFall~\cite{re48_UP-Fall}. In this study, FallAllD and SisFall datasets were chosen to validate the proposed MLMEC framework and the use of the KD model for managing the trade-off between accuracy and latency in designing FD systems. These two datasets are closer to real-life fall types and an ADL than other datasets, and they have common detection sites that can be used for reference and comparison. The datasets use accelerometers with a sampling rate of 200 Hz, which is helpful for the model performance.

\subsubsection{FallAllD Dataset}
M. Saleh~\textit{et al.}~\cite{re25_FallAllD} proposed a public dataset that involved placing an inertial measurement unit (IMU) on the neck, chest, and waist to capture various types of falls during both non-activity and activity. Each IMU included a triaxial accelerometer, gyroscope, magnetometer, and barometer. Fifteen healthy young participants took part in the study, providing a total of 1053 ADLs and 425 fall trials. Accelerometer data from the waist was the focus due to its prominence and ease of distinction. Tab.~\ref{tb_FallAllD} shows the details of the FallAllD dataset. The FallAllD dataset comprised 15 healthy young participants, encompassing eight males and seven females, aged between 21 and 53 years, weighing between 48 and 85 kg, and measuring in height from 158 to 187 cm. The average age, height, and weight were 32 years, 171 cm, and 67 kg, respectively.

\begin{table}[htbp]
\centering
\caption{The information on participation in the FallAllD dataset}
\scalebox{0.9}{
      \begin{tabular}{|c|c|c|c|}
      \hline
        \textbf{Sex} & \textbf{Age (years)} & \textbf{Height(m)} & \textbf{Weight(kg)} \\
        \hline
        \textbf{8 Males}  & \multirow{2}{*}{21-53} & \multirow{2}{*}{1.58-1.87} & \multirow{2}{*}{48-85 }
        \\\cline{1-1}
        \textbf{7 Females}& & & \\\cline{1-4}
       \end{tabular}}
     \label{tb_FallAllD}
\end{table}

\subsubsection{SisFall Dataset}
A. Sucerquia~\textit{et al.}~\cite{re26_SisFall} combined data from two age groups: 23 young participants and 14 participants aged 62 or older. This study used data from 21 young participants, excluding the older group without fall events. Two young participants were excluded due to incomplete ADL trials. The data, collected using a self-developed IMU fixed at the waist, includes 1575 falls and 1659 ADL trials, sampled at 200 Hz. Tab.~\ref{tb_SisFall} shows the details of the SisFall dataset. The 21 young participants consisted of 10 males and 11 females, with ages ranging from 19 to 30 years. The weight of males varied from 58 to 81 kg, while females weighed between 42 and 63 kg. Heights ranged from 165 to 183 cm for males and 149 to 169 cm for females. On average, the participants were 25 years old, had a height of 165 cm, and weighed 57.7 kg.

\begin{table}[htbp]
\centering
\caption{The information on participation in the SisFall dataset}
\scalebox{0.9}{
      \begin{tabular}{|c|c|c|c|}
      \hline
        \textbf{Sex} & \textbf{Age (years)} & \textbf{Height(m)} & \textbf{Weight(kg)} \\
        \hline
        \textbf{10 Males}  & \multirow{2}{*}{19-30} & 1.65-1.83 & 58-81 
        \\\cline{1-1}\cline{3-4}
        \textbf{11 Females} &  & 1.49-1.69 & 42-63\\\hline
       \end{tabular}}       
     \label{tb_SisFall}
\end{table}

\subsection{Performance Evaluations of MLMEC Systems}
\label{sec:Performance evaluations of MLMEC systems}

In this subsection, we compare the performances of dual- and triple-layer MECs FD systems with and without the KD approach.
The candidate models of each MEC and CC include ResNet18, ResNet50, ResNet101, MobileNetV3, and basic CNN models. 
The hyperparameter settings of the ResNet series and MobileNetV3 are the same as~\cite{re46_ResNet} and~\cite{re45_MobileNetV3}, respectively.
The hyperparameter details of the CNN model were as follows: number of filters = 64, filter width = 1, epochs = 200, batch size = 64, and learning rate = 0.001.
Further details of the parameter settings can be found on GitHub$^{1}$.
The results of all models are averaged over ten experiments.
In terms of the loss function, we set $\lambda$ = 0.5 and $T$ = 20. 

Fig.~\ref{fig6:Comparison of the MEC, MLMEC, and MLMEC with KD architectures} presents the performances in terms of $ACC$, $REC$, $PRE$, and $F1$. 
As shown in Fig.~\ref{fig6:Comparison of the MEC, MLMEC, and MLMEC with KD architectures}(a) and Fig.~\ref{fig6:Comparison of the MEC, MLMEC, and MLMEC with KD architectures}(b), compared to the dual-layer MECs, the performance of triple-layer MECs shows improvements in $ACC$ and $REC$, while $PRE$ and $F1$ scores are reduced slightly on the FallAllD and SisFall datasets. 
The inclusion of weaker-performing DL models in the MLMEC system negatively impacts the overall performance. 
It is worth mentioning that Fig.~\ref{fig6:Comparison of the MEC, MLMEC, and MLMEC with KD architectures}(a) and Fig.~\ref{fig6:Comparison of the MEC, MLMEC, and MLMEC with KD architectures}(b) show that all evaluation metrics are effectively enhanced using the KD approach for both FallAllD and SisFall datasets.

\begin{figure}[h]
    \subfloat[\label{subfig:6a}]{%
      \includegraphics[width=0.45\textwidth]{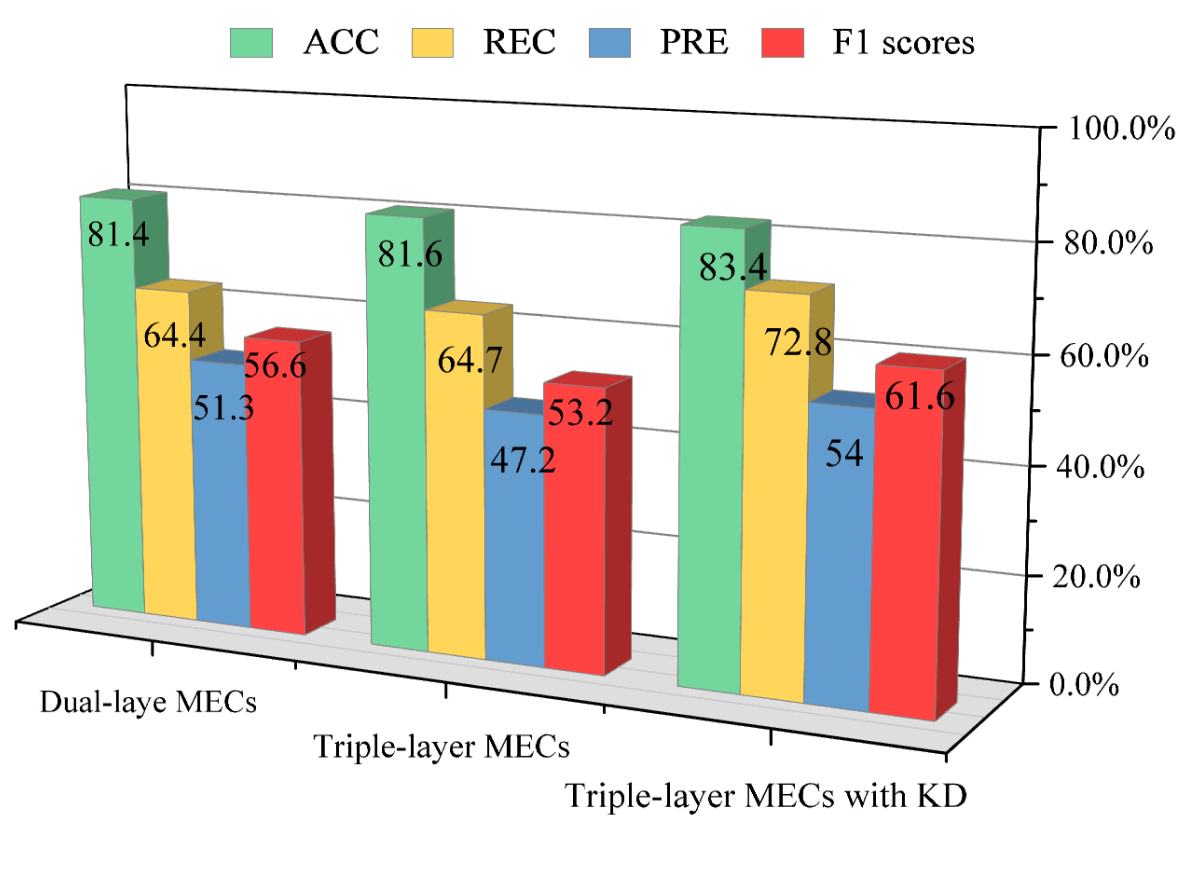}
    }
    \hfill
    \subfloat[\label{subfig:6b}]{%
      \includegraphics[width=0.45\textwidth]{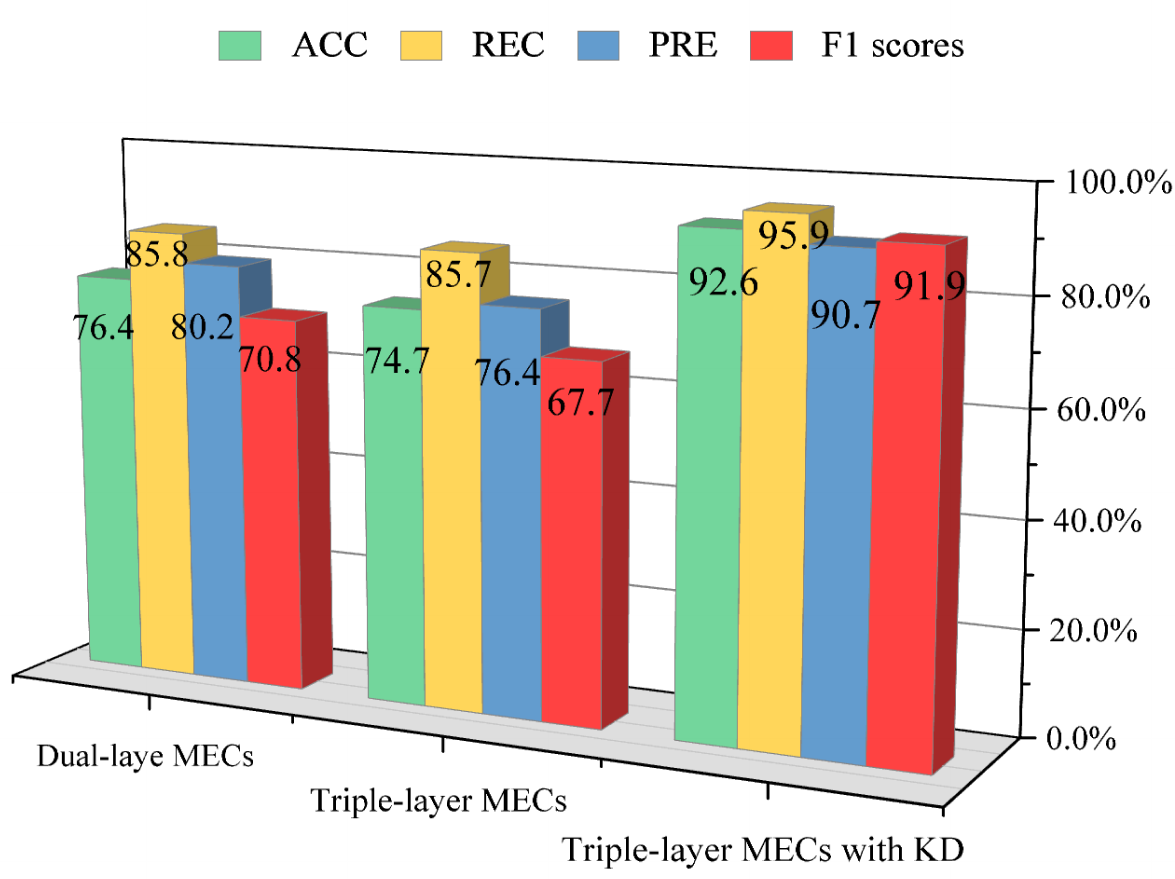}
    }
   \caption{Comparison of the dual-layer MECs, triple-layer MECs, and triple-layer MECs with KD architectures using two datasets (a) FallAllD (b) SisFall.}
\label{fig6:Comparison of the MEC, MLMEC, and MLMEC with KD architectures}
\end{figure}

\begin{figure}[htbp]
    \subfloat[\label{subfig:7a}]{%
      \includegraphics[width=0.45\textwidth]{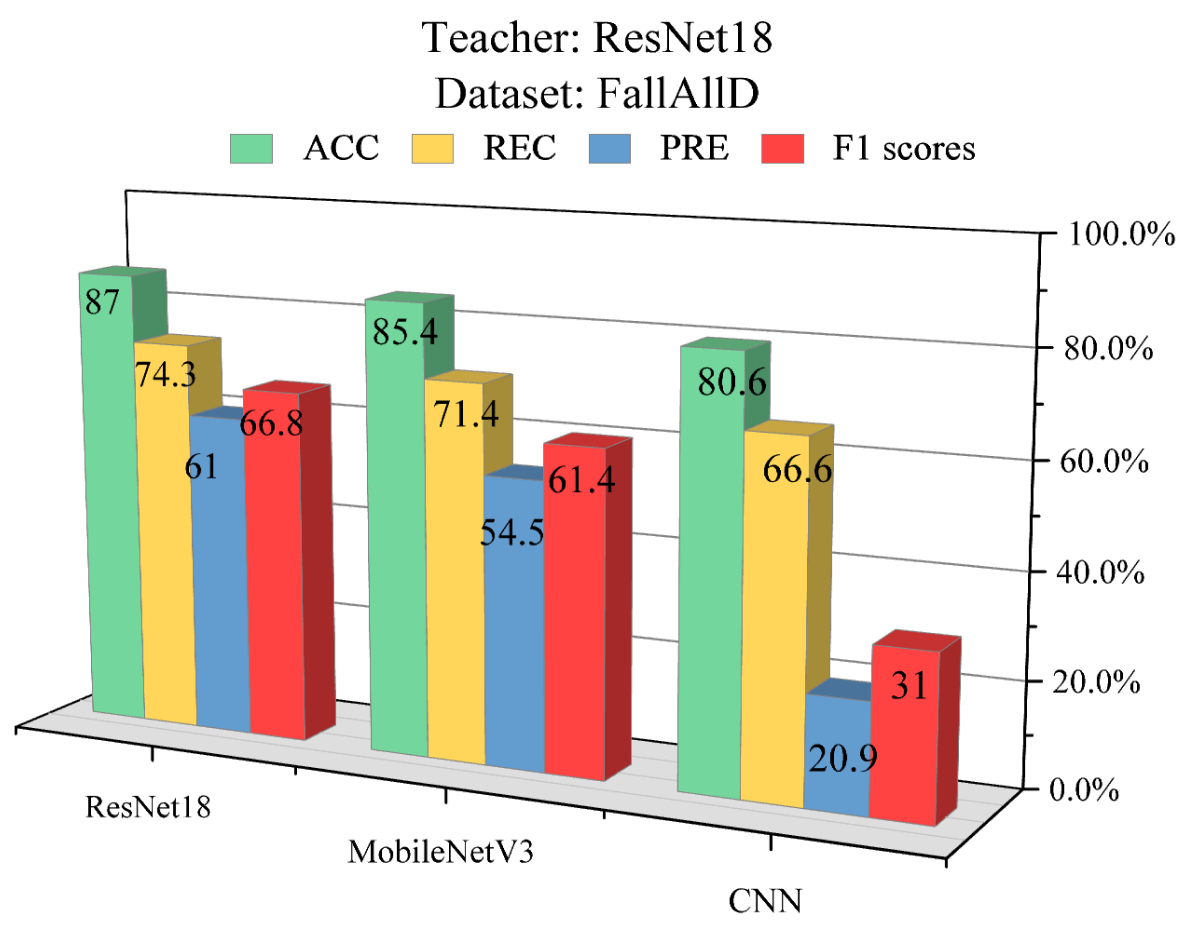}
    }
    \hfill
    \subfloat[\label{subfig:7b}]{%
      \includegraphics[width=0.45\textwidth]{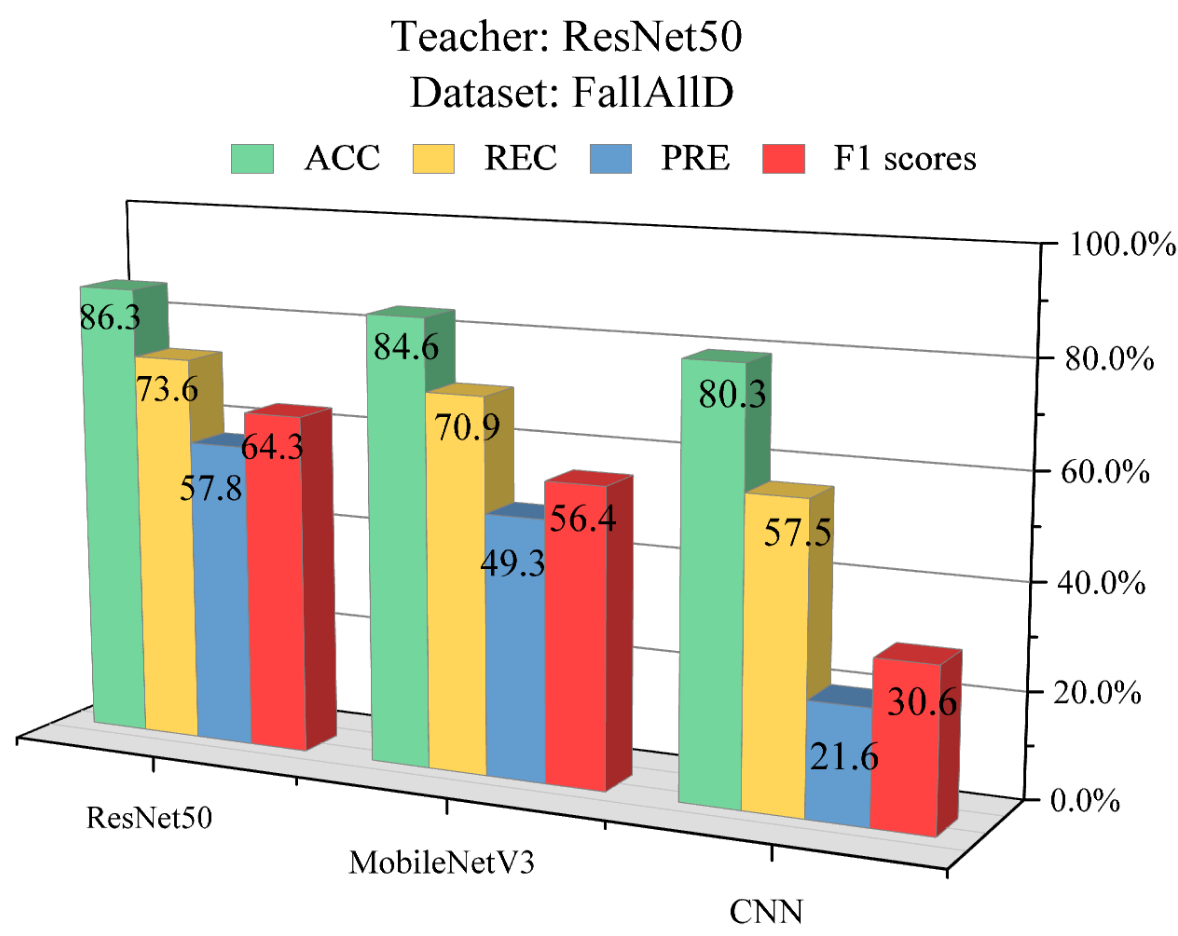}
    }
    \hfill
    \subfloat[\label{subfig:7c}]{%
      \includegraphics[width=0.45\textwidth]{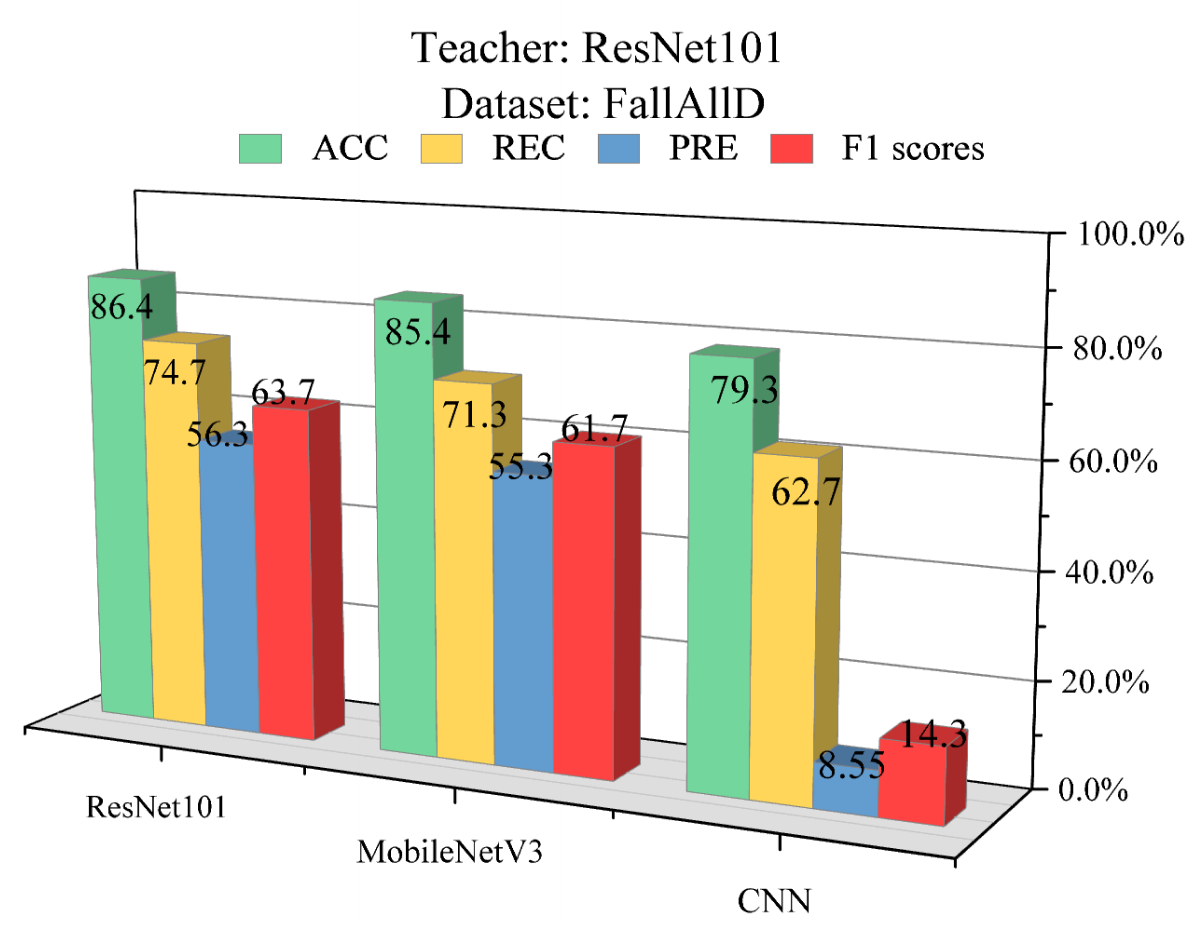}
    }
   \caption{The different combinations for the dual-layer with KD based on the FallAllD dataset (a) Teacher model ResNet18. (b) Teacher model ResNet50. (c) Teacher model ResNet101.}
\label{fig7:dual-layer KD based on the FallAllD dataset}
\end{figure}

\begin{figure}[htbp]
    \subfloat[\label{subfig:8a}]{%
      \includegraphics[width=0.45\textwidth]{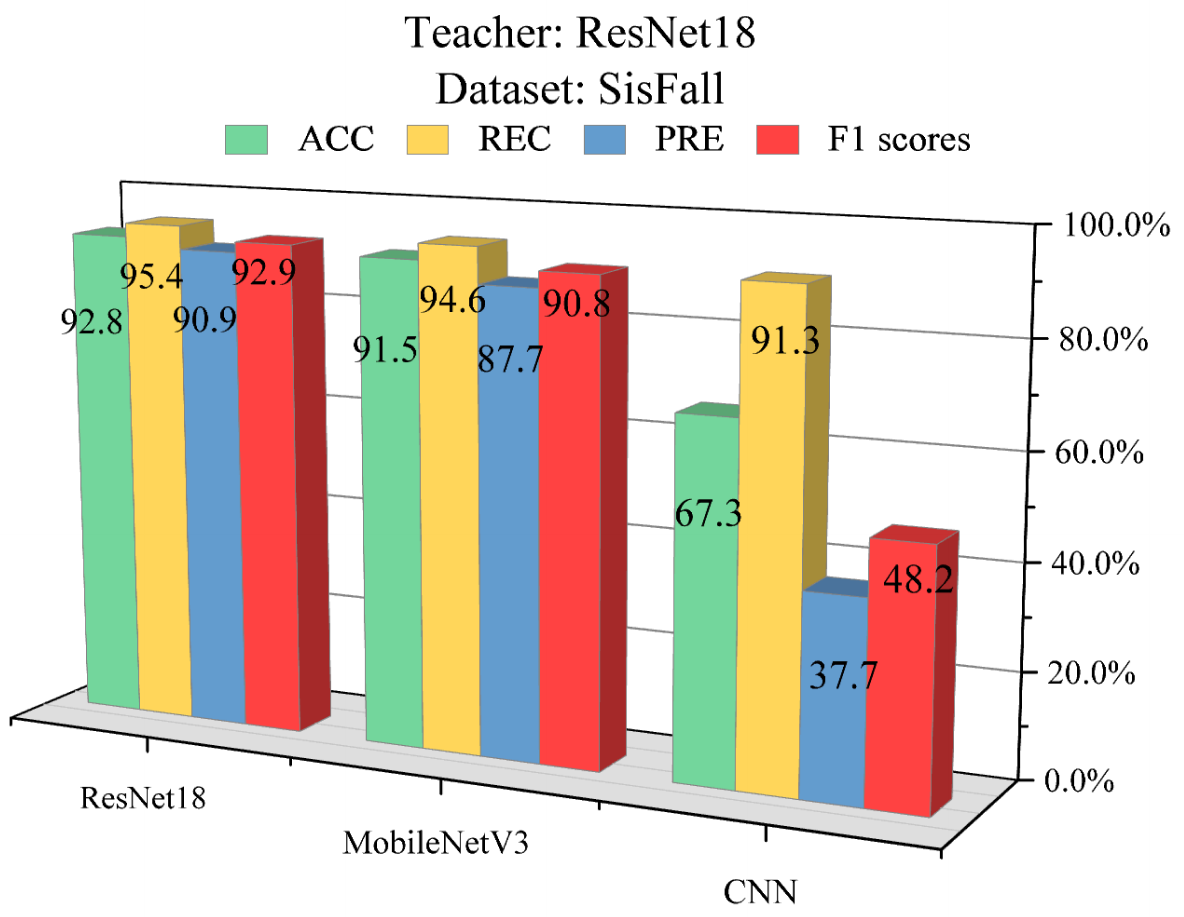}
    }
    \hfill
    \subfloat[\label{subfig:8b}]{%
      \includegraphics[width=0.45\textwidth]{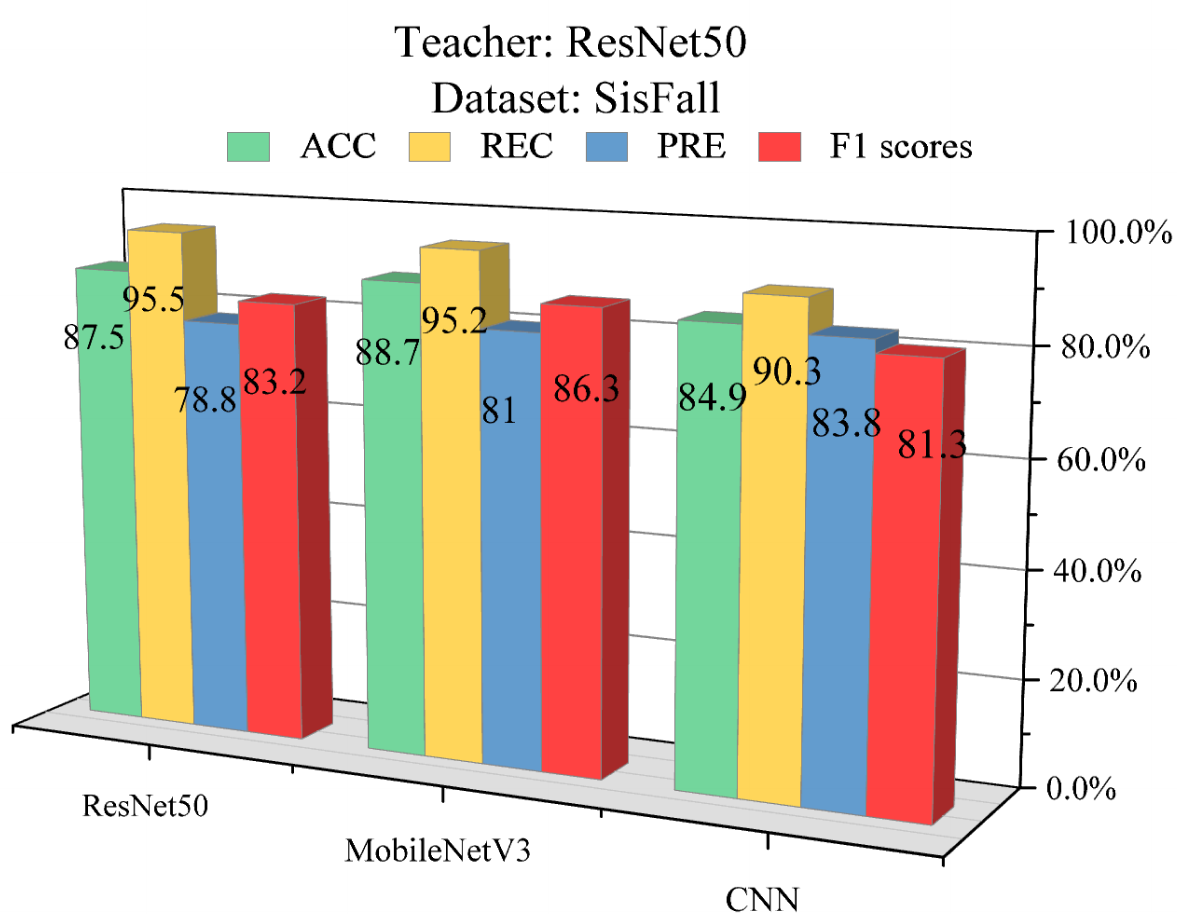}
    }
    \hfill
    \subfloat[\label{subfig:8c}]{%
      \includegraphics[width=0.45\textwidth]{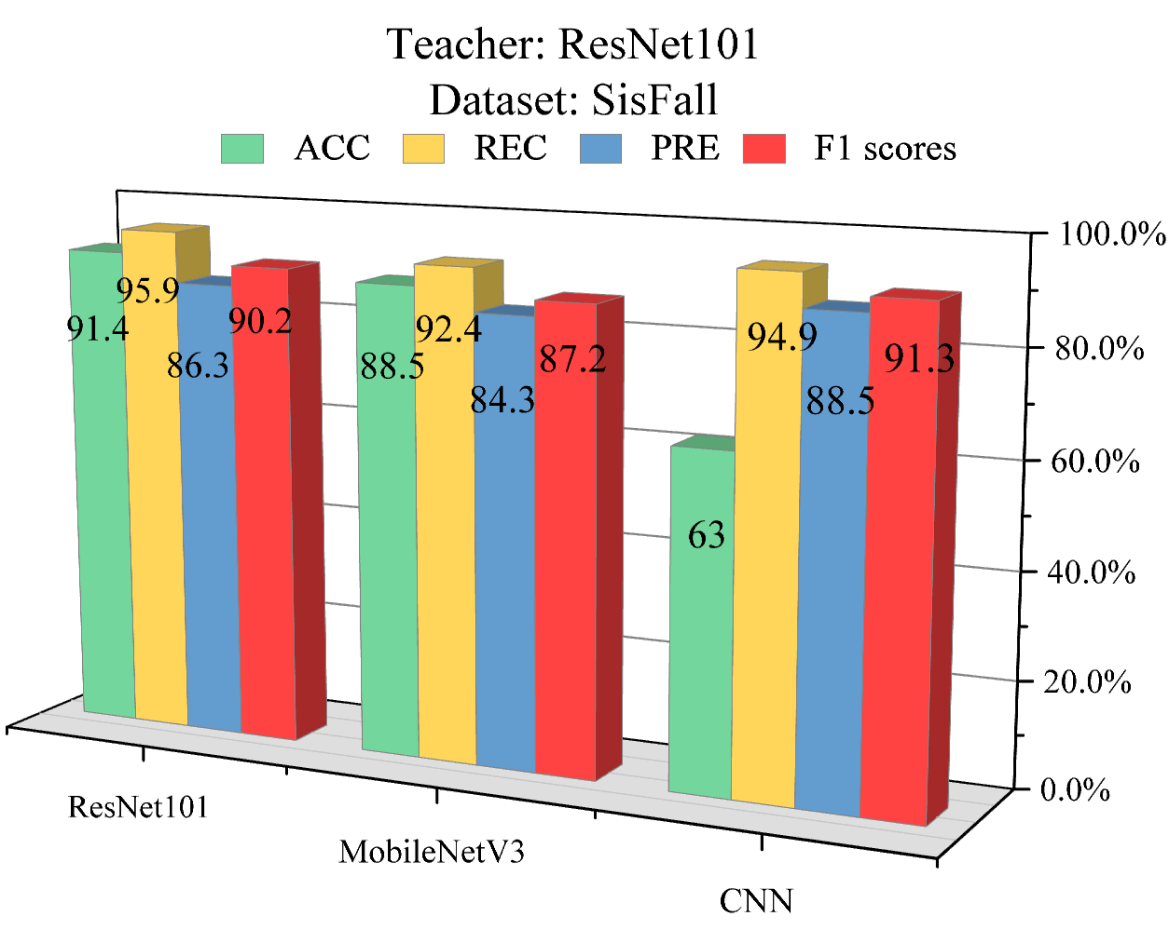}
    }
   \caption{The different combinations for the dual-layer with KD based on the SisFall dataset (a) Teacher model ResNet18. (b) Teacher model ResNet50. (c) Teacher model ResNet101.}
\label{fig8:dual-layer KD based on the SisFall dataset}
\end{figure}

Here, our goal is to find outstanding teacher and TA models for triple-layer MECs with the KD approach. Therefore, we first perform the experiments of dual-layer MECs to observe the results of the different student models in combination with the same teacher model.
Fig.~\ref{fig7:dual-layer KD based on the FallAllD dataset} to \ref{fig8:dual-layer KD based on the SisFall dataset} present the performance of the different student models with the KD approach based on the same teacher model. %where (O) represents the original student model, and (D) represents the distilled student model.

As shown in Fig.~\ref{fig7:dual-layer KD based on the FallAllD dataset}(a), ResNet18 is adopted as the teacher model and tested on the FallAllD dataset. We adopt ResNet18, MobileNetV3, and CNN as the student models with the KD approach.
ResNet18 with KD outperformed the other models.
Similarly, the same phenomenon is demonstrated in Fig.~\ref{fig7:dual-layer KD based on the FallAllD dataset}(b) and~\ref{fig7:dual-layer KD based on the FallAllD dataset}(c).
As shown in Fig.~\ref{fig7:dual-layer KD based on the FallAllD dataset}(b) and~\ref{fig7:dual-layer KD based on the FallAllD dataset}(c), the performance of ResNet50 and ResNet101 with KD outperforms the other models, respectively.

As shown in Fig.~\ref{fig8:dual-layer KD based on the SisFall dataset}(a), ResNet18 is adopted as the teacher model and tested on the SisFall dataset. We also adopted ResNet18 and MobileNetV3 as the student models with the KD approach.
Similarly, the performance of ResNet18 outperformed the other models, while MobileNetV3 performed as the sub-optimal student model.
As shown in Fig.~\ref{fig8:dual-layer KD based on the SisFall dataset}(b), the performance of MobileNetV3 was slightly better than that of ResNet50 and CNN. 
As shown in Fig.~\ref{fig8:dual-layer KD based on the SisFall dataset}(c), the performance of ResNet101 is slightly better than MobileNetV3 and CNN.
In addition, it is worth noting that increasing the depth of the ResNet teacher model may not necessarily improve the performance of the student model and could even lead to slight degradation. Please note that compared to the ResNet family, MobileNet V3 has a much simpler architecture and requires fewer computational resources during inference.
Therefore, selecting ResNet18 and MobileNetV3 as the teacher and TA models, respectively, is a priority option to consider.

\begin{figure}[h]
    \subfloat[\label{subfig:9a}]{%
      \includegraphics[width=0.45\textwidth]{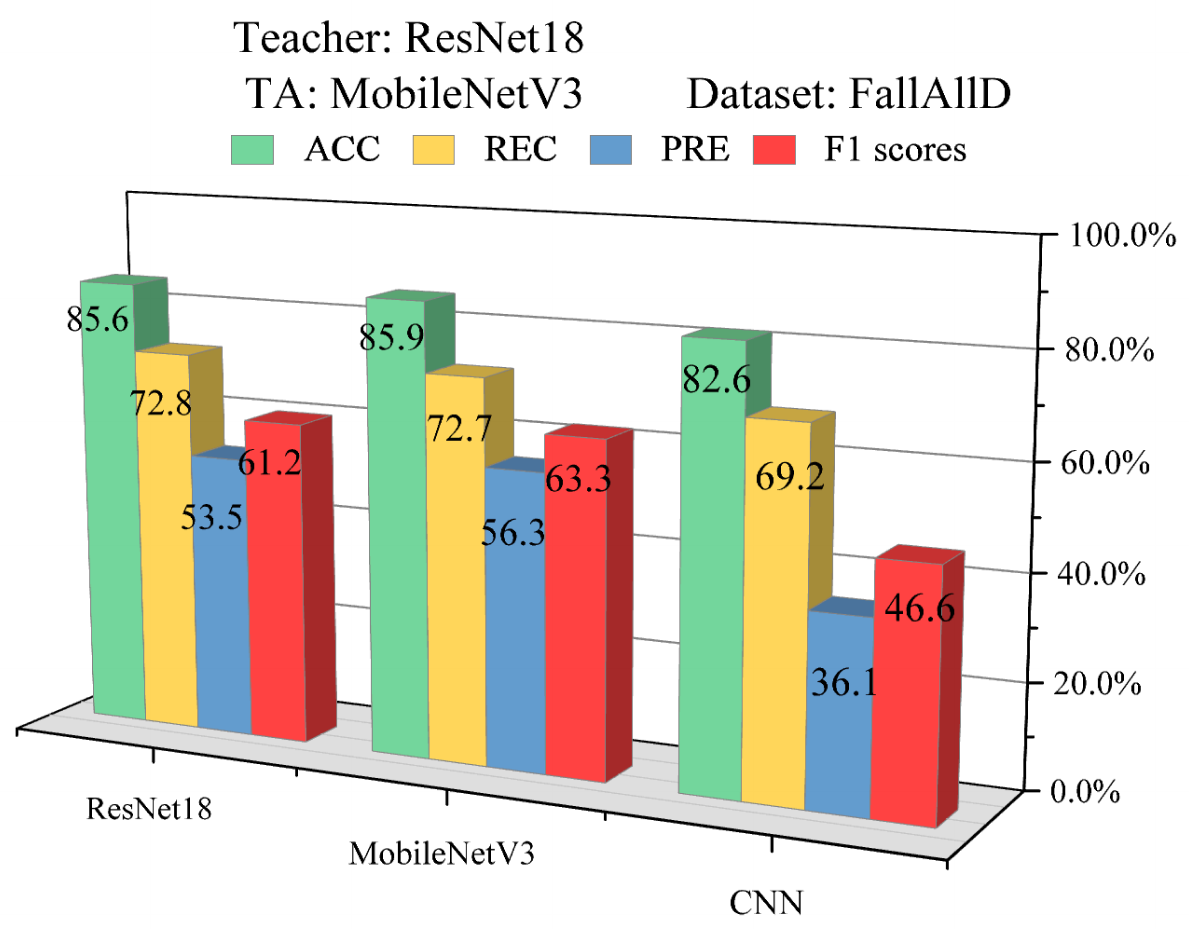}
    }
    \hfill
    \subfloat[\label{subfig:9b}]{%
      \includegraphics[width=0.45\textwidth]{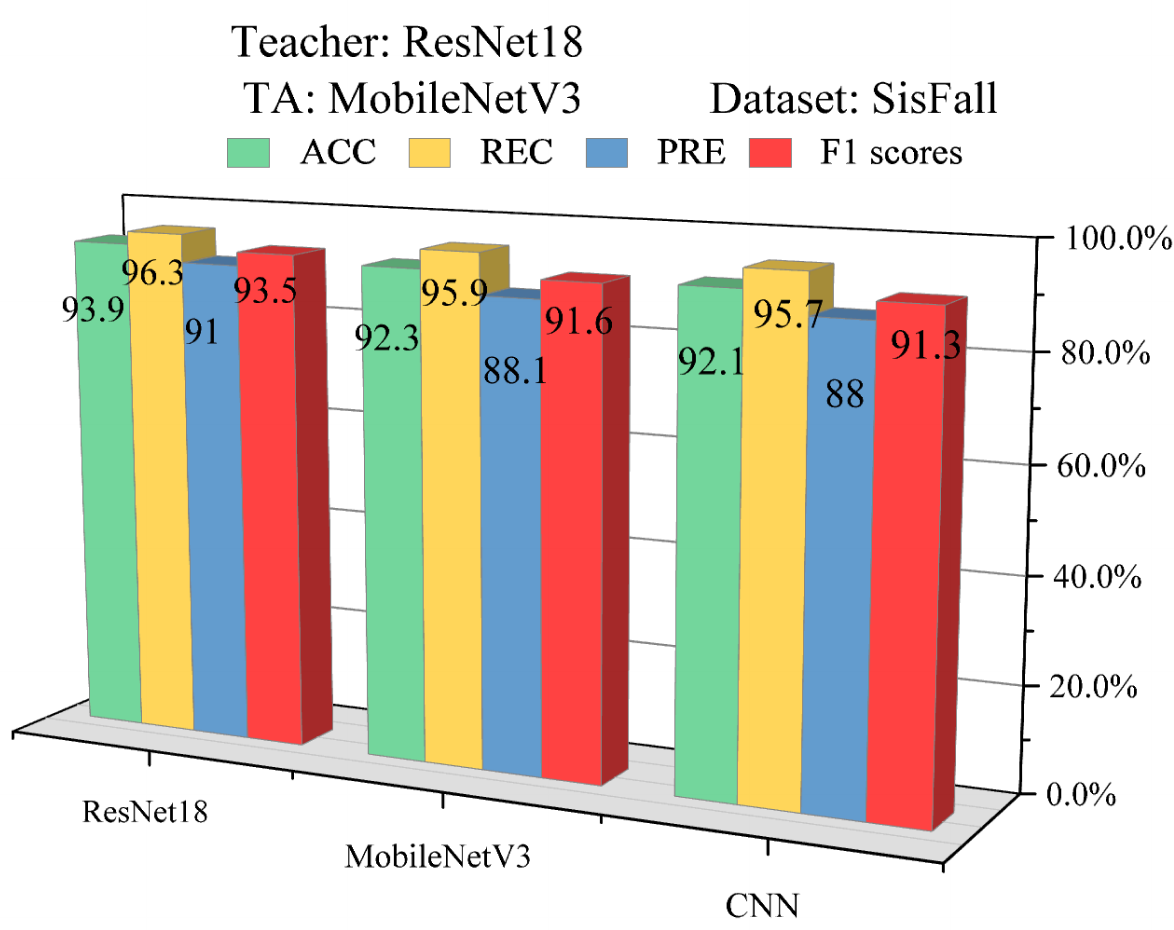}
    }

   \caption{The different combinations for the triple-layer with KD (a) FallAllD dataset. (b) Sisfall dataset.}
\label{fig9:Comparison between the triple-layer KD}
\end{figure}

As shown in Fig.~\ref{fig9:Comparison between the triple-layer KD}, using the FallAllD dataset, ResNet18, and MobileNetV3, respectively, were adopted as the teacher and TA models.
We adopt ResNet18, MobileNetV3, and CNN as the candidate student models with the KD approach.
As shown in Fig.~\ref{fig9:Comparison between the triple-layer KD}(a), the performance of the ResNet18 student model in the triple-layer is not superior to dual-layer in Fig.~\ref{fig7:dual-layer KD based on the FallAllD dataset}(a). 
As shown in Fig.~\ref{fig9:Comparison between the triple-layer KD}(b), the performance of triple-layer KD consistently outperforms the dual-layer KD by using the SisFall dataset.

\subsection{Comparison of the Teacher Models}
\label{subsec:Model KD model ranking selection}
In this subsection, our goal is to further compare the performance of using different teacher models from ResNet, MobileNetV3, and CNN.
For comparing the different DL models, the calculations of $ACC_{imp}$, $REC_{imp}$, $PRE_{imp}$, and $F1_{imp}$ are adopted as shown
\begin{align}
\tiny
     ACC_{imp}=
    \frac{(\mathrm{Distilled}\ ACC-\mathrm{Original}\ ACC)}{\mathrm{Original}\ ACC} \times 100\%,
        \label{eq26:ACC}
\end{align}

\begin{align}
\tiny
    REC_{imp}=
    \frac{(\mathrm{Distilled}\ REC-\mathrm{Original}\ REC)}{\mathrm{Original}\ REC} \times 100\%,
        \label{eq27:REC}     
\end{align}

\begin{align}
\tiny
       PRE_{imp}=
       \frac{(\mathrm{Distilled}\ PRE-\mathrm{Original}\ PRE)}{\mathrm{Original}\ PRE} \times 100\%,
        \label{eq28:PRE}
\end{align}

\begin{align}
\tiny
       F1_{imp}=
       \frac{(\mathrm{Distilled}\ F1-\mathrm{Original}\ F1)}{\mathrm{Original}\ F1} \times 100\%.
        \label{eq29:F1}       
\end{align}
Tables~\ref{tb4_COMPARISON CNN MOBILENETV3} and~\ref{tb5_COMPARISON ResNet} compare the performances of different teacher models using the KD approach in the dual-layer framework.
As shown in Tables~\ref{tb4_COMPARISON CNN MOBILENETV3} and~\ref{tb5_COMPARISON ResNet}, the distillation effect of using lighter CNN models and MobileNetV3 as the teacher models was significantly worse than that of ResNet as the teacher. 
Therefore, ResNet was suitably adopted as the teacher model for the CC.  
These results demonstrate that the various depths of the ResNet model exhibit no obvious improvement in the KD approach. 
Even when a deeper ResNet is adopted, the teacher model may not improve the performance of the student model and may even lead to slight degradation.
Hence, ResNet18 is adopted as the teacher model in the proposed MLMEC framework. 
It is worth noticing that the highest improvement rate was achieved when ResNet and MobileNetV3 served as teacher and student models, respectively.
However, the primary goal of the proposed MLMEC system is to enhance the lighter model positioned in the lower-layer MEC, rather than solely optimizing the accuracy performance. 
Thus, we adopted MobileNetV3 as the TA model in the triple-layer MECs framework.

\begin{table}[htbp]
\centering
\caption{Comparison of CNN and MobileNetV3.}
\scalebox{0.8}{
      \begin{tabular}{|c|c|c|c|c|c|}
      \hline
        \textbf{Teacher Model} & \textbf{Student Model} &\textbf{$ACC_{imp}$}&\textbf{$REC_{imp}$}&\textbf{$PRC_{imp}$}&\textbf{$F1_{imp}$}\\
        \hline
        \textbf{\multirow{4}{*}{MobileNetV3}}  & \textbf{CNN} & 0.29\% & 0.62\% & 3.38\% & 2.69\% \\\cline{2-6}
        &\textbf{ResNet18} & \textbf{0.94\%} & 1.38\% &3.17\% & 2.81\% \\\cline{2-6}
        & \textbf{ResNet50} & -2.21\% & 0.78\% & 1.96\% & -2.47\% \\\cline{2-6}
        & \textbf{ResNet101} &  0.13\% & 0.01\% & 3.71\% & 2.48\% \\\hline
        \textbf{\multirow{4}{*}{CNN}} & \textbf{MobileNetV3} & 0.82\% & 1.27\% &\textbf{3.84\%} & \textbf{2.93\%} \\\cline{2-6}
        & \textbf{ResNet18} & -0.07\% & -2.88\% &-11.10\% & -13.71\% \\\cline{2-6}
        & \textbf{ResNet50} & 0.37\% & \textbf{1.81\%} & -4.16\% & -2.97\% \\\cline{2-6}
        & \textbf{ResNet101} &  -2.69\% & 0.33\% & -4.15\% & -10.12\% \\\hline      
        \end{tabular}}
     \label{tb4_COMPARISON CNN MOBILENETV3}
\end{table}

\begin{table}[htbp]
\centering
\caption{Comparison of ResNet18, ResNet50, and ResNet101.}
\scalebox{0.8}{
      \begin{tabular}{|c|c|c|c|c|c|}
      \hline
        \textbf{Teacher Model} & \textbf{Student Model} &\textbf{$ACC_{imp}$}&\textbf{$REC_{imp}$}&\textbf{$PRC_{imp}$}&\textbf{$F1_{imp}$}\\
        \hline
        \textbf{\multirow{3}{*}{ResNet18}}  & \textbf{CNN} & 0.20\% & 11.45\% &-32.79\% & -23.20\% \\\cline{2-6}
        & \textbf{ResNet18} &  4.96\% & 11.64\% & 43.13\% & 29.07\% \\\cline{2-6}
        & \textbf{MobileNetV3} &  4.56\% & 4.36\% & \textbf{92.17\%} & \textbf{53.54\%} \\\hline
        \textbf{\multirow{3}{*}{ResNet50}}  & \textbf{CNN} & -0.57\% & -4.95\% &-33.31\% & -26.94\% \\\cline{2-6}
        & \textbf{ResNet50} &  5.04\% & 11.00\% & 59.09\% & 37.63\% \\\cline{2-6}
        & \textbf{MobileNetV3} &  3.73\% & 4.62\% & 76.34\% & 42.52\% \\\hline
        \textbf{\multirow{3}{*}{ResNet101}}  & \textbf{CNN} & -1.55\% & 4.07\% &-71.65\% & -63.62\% \\\cline{2-6}
        & \textbf{ResNet101} &  \textbf{5.73\%} & \textbf{12.72\%} & 77.45\% & 49.23\% \\\cline{2-6}
        & \textbf{MobileNetV3} &  4.46\% & 4.36\% & 87.99\% & 50.13\% \\\hline       
        \end{tabular}}
     \label{tb5_COMPARISON ResNet}
\end{table}

\subsection{Comparison of the Student Models}
\label{Comparing the different MLMEC models with KD.}
In this section, our goal is to compare student models considering low computational complexity and good detection rates from ResNet, MobileNetV3, and CNN.
We investigated the computational complexity of each layer in terms of floating-point operation counts (FLOPs), parameter number, and accuracy using the FallAllD and SisFall datasets as shown in Table~\ref{tb6_MLMEC FRAMEWORK  FALLALLD KD} and~\ref{tb7_MLMEC FRAMEWORK SisFall KD}, respectively. 
FLOPs and parameter calculations are used to evaluate the computational complexity of DL models in~\cite{re53_Embedded,re54_Pruning,re55_Mobile-Former}.
To compute the number of floating-point operations (FLOPs), we adopt the THOP package of Pytorch. The source code for the package can be found on GitHub\footnote{The source code is available at https://github.com/Lyken17/pytorch-OpCounter} and contains comprehensive details about parameter settings. For example, FLOPs of fully connected layers are computed as
\begin{equation}
        \mathrm{FLOPs}_{FC}=(2I-1)O,
        \label{eq30:Flopsfc}       
\end{equation}
where $I$ and $O$ are the dimensions of input and output, respectively. FLOPs of convolution layers are calculated as
\begin{equation}
        \mathrm{FLOPs}_{Conv}=2HW(C_{in}K^{2}+1)C_{out},
        \label{eq31:Flopsconv}       
\end{equation}
where $H$, $W$, and $C_{in}$ are the height, width, and number of channels of the input feature map, $K$ is the kernel width, and $C_{out}$ is the number of output channels. The authors in ~\cite{re55_Mobile-Former} compared the FLOPs of ResNet18 and MobileNetV3.
The KD approach does not change the computational load of the model. We present FLOPs to thoroughly illustrate the computational complexity of each model and the relative size of the models placed in each layer of the proposed MLMEC FD system.

To determine the amount of data in each layer, we used threshold-based upward judgment to determine the uncertain range of whether the data should be sent to the upper-layer model for further detection. 
In the proposed MLMEC, we adopted ResNet18, MobileNetV3, and CNN as CC, MEC 2, and MEC 1, respectively.

\begin{table}[htbp]
\centering
\caption{Comparison of Computational Complexity using FallAllD.}
\scalebox{0.8}{
      \begin{tabular}{|c|c|c|c|}
      \hline
        &\textbf{MEC 1 CNN} & \textbf{MEC 2 MobileNetV3} &\textbf{CC ResNet18}\\\hline
        \textbf{FLOPs} & 18.17K & 85.87M & 1.292G\\\hline
        \textbf{Parameters} & 0.058K & 1.237M & 13.94M\\\hline
        \textbf{Accuracy} & 82.59\% & 85.92\% & 85.57\%\\\hline
        \end{tabular}}
     \label{tb6_MLMEC FRAMEWORK  FALLALLD KD}
\end{table}

\begin{table}[htbp]
\centering
\caption{Comparison of Computational Complexity using SisFall.}
\scalebox{0.8}{
      \begin{tabular}{|c|c|c|c|}
      \hline
        &\textbf{MEC 1 CNN} & \textbf{MEC 2 MobileNetV3} &\textbf{CC ResNet18}\\\hline
        \textbf{FLOPs} & 18.17K & 85.87M & 1.292G\\\hline
        \textbf{Parameters} & 0.058K & 1.237M & 13.94M\\\hline
        \textbf{Accuracy} & 89.15\% & 92.88\% & 93.89\%\\\hline
        \end{tabular}}
     \label{tb7_MLMEC FRAMEWORK SisFall KD}
\end{table}

To ensure the applicability of uncertain data in real-world scenarios, the concept of a threshold is proposed during the testing phase, utilizing the softmax output judgment process. 
The data were categorized into three groups: fall types, ADLs, and uncertain data. 
If the detection of the current MEC layer is classified as uncertain, it is sent to the next MEC layer. 
For example, if the softmax output of the current layer is larger than the maximum threshold value (0.8 in this study), it represents a fall type. Conversely, if the output falls below the minimum threshold value (0.2 in this study), it represents ADLs.

Table~\ref{tb8_Upward-THRESHOLD TESTING FALLALLD} and Table \ref{tb9_Upward-THRESHOLD TESTING Sisfall} present the data amount of each layer in the MLMEC under different threshold settings by using the FallAllD and Sisfall datasets. The FallAllD dataset contains 1,428,600 testing data points, while the SisFall dataset has 2,463,200 testing data points.

\begin{table}[htbp]
\centering
\caption{Threshold-based for uncertain data using FallAllD.}
\scalebox{0.8}{
      \begin{tabular}{|c|c|c|c|c|}
      \hline
       \textbf{\makecell{Maximum\\ Back-ward\\ Threshold}} & \textbf{\makecell{Minimum\\Back-ward \\Threshold}} & \textbf{\makecell{MEC 1 \\CNN \\(\# of data)}} &\textbf{\makecell{MEC 2 \\MobileNetV3\\ (\# of data)}} & \textbf{\makecell{CC\\ResNet18\\(\# of data)}}\\\hline
        \textbf{0.6} & \textbf{0.4} & 1373600 & 55000 & 0\\\hline
        \textbf{0.7} & \textbf{0.3} & 1315600 & 91400 & 21600\\\hline
        \textbf{0.8} & \textbf{0.2} & 1233000 & 148400 & 47200\\\hline
        \textbf{0.9} & \textbf{0.1} & 1102400 & 211800 & 114400\\\hline
        \end{tabular}}
     \label{tb8_Upward-THRESHOLD TESTING FALLALLD}
\end{table}

\begin{table}[htbp]
\centering
\caption{Threshold-based for uncertain data using SisFall.}
\scalebox{0.8}{
      \begin{tabular}{|c|c|c|c|c|}
      \hline
       \textbf{\makecell{Maximum\\ Back-ward\\ Threshold}} & \textbf{\makecell{Minimum\\Back-ward \\Threshold}} & \textbf{\makecell{MEC 1 \\CNN \\(\# of data)}} &\textbf{\makecell{MEC 2 \\MobileNetV3\\ (\# of data)}} & \textbf{\makecell{CC\\ResNet18\\(\# of data)}}\\\hline
        \textbf{0.6} & \textbf{0.4} & 2270800 & 171600 & 20800\\\hline
        \textbf{0.7} & \textbf{0.3} & 2057600 & 254800 & 150800\\\hline
        \textbf{0.8} & \textbf{0.2} & 1813200 & 215200 & 434800\\\hline
        \textbf{0.9} & \textbf{0.1} & 868400 & 378400 & 1216400\\\hline
        \end{tabular}}
     \label{tb9_Upward-THRESHOLD TESTING Sisfall}
\end{table}

 %Based on the results shown in Tables~\ref{tb10 FALLAIID} and~\ref{tb11 SisFall}, these results are primarily aimed at achieving a closer approximation to real-world applications. 

%In this experiment, eq. 11$T_{fall}^{Max}$ and $T_{fall}^{Min}$ are set to 0.8 and 0.2, respectively, eq. 10 $T$ is set to 20.
The preset threshold values were chosen to maximize the resource efficiency of the MECs.
Based on the threshold-based judgment for uncertain data, the amounts of data in each layer were compared, as shown in Tables \ref{tb12_TWO-TIER ARCHITECTURE}, \ref{tb13_THE MLMEC ARCHITECTURE}, and \ref{tb14_The MLMEC architecture with KD results}. 
In addition to comparing the amounts of processed data, the latency of each layer was calculated. 
The parameters $s_{n}^{i}$ Eq.~\ref{eq25:Ln} represent the data transmission rates per second and the parameters $\theta_{n}^{i}$ denote the CPU computation loads.

\begin{table}[t]
\centering
\caption{The computational complexity in dual-layer MEC without KD.}
\scalebox{0.8}{
      \begin{tabular}{|c|c|c|c|}
      \hline
       \textbf{Data} & \textbf{\makecell{MEC 1 \\CNN \\(\# of data)}} & \textbf{\makecell{CC\\ResNet18\\(\# of data)}}& \textbf{Latency(ms)}\\\hline
        \textbf{FallAllD} & 926400 & 502200 & 2145.12 \\\hline
        \textbf{SisFall} & 1813200 & 650000 & 2752.54 \\\hline
        \end{tabular}}
     \label{tb12_TWO-TIER ARCHITECTURE}
\end{table}

\begin{table}[t]
\centering
\caption{The computational complexity in triple-layer MEC without KD.}
\scalebox{0.8}{
      \begin{tabular}{|c|c|c|c|c|c|}
      \hline
       \textbf{Data} & \textbf{\makecell{MEC 1 \\CNN \\(\# of data)}} & \textbf{\makecell{MEC 2 \\MobileNetV3 \\(\# of data)}}
       &\textbf{\makecell{CC\\ResNet18\\(\# of data)}}&\textbf{\makecell{MEC 1 to\\MEC 2\\Latency\\ (ms)}} &\textbf{\makecell{MEC 2 to\\CC\\Latency\\ (ms)}}\\\hline
        \textbf{FallAllD} & 926400 & 430200 & 72000 & 2045.71 & 344.48 \\\hline
        \textbf{SisFall} & 1813200 & 215200 & 434800 & 2691.28 & 1841.93 \\\hline
        \end{tabular}}
     \label{tb13_THE MLMEC ARCHITECTURE}
\end{table}

The latencies of the dual-layer and triple-layer MECs architecture without KD are presented in Table~\ref{tb12_TWO-TIER ARCHITECTURE} and Table~\ref{tb13_THE MLMEC ARCHITECTURE}, respectively. 
Since the dual-layer only has the latency from MEC 1 to CC, the latency from MEC 1 to MEC 2 of the triple-layer is used to compare. The improvement in latency is approximately 4.64\%.
 
\begin{table}[t]
\centering
\caption{The computational complexity in triple-layer MEC with KD.}
\scalebox{0.8}{
      \begin{tabular}{|c|c|c|c|c|c|}
      \hline
       \textbf{Data} & \textbf{\makecell{MEC 1 \\CNN\\(\# of data)}} & \textbf{\makecell{MEC 2 \\MobileNetV3 \\(\# of data)}}
       &\textbf{\makecell{CC\\ResNet18\\(\# of data)}}&\textbf{\makecell{MEC 1 to\\MEC 2\\Latency\\ (ms)}} &\textbf{\makecell{MEC 2 to\\CC\\Latency\\ (ms)}}\\\hline
        \textbf{FallAllD} & 1233000 & 148400 & 47200 & 844.86 & 250.92 \\\hline
        \textbf{SisFall} & 2012400 & 322800 & 128000 & 1855.35 & 562.11 \\\hline
        \end{tabular}}
     \label{tb14_The MLMEC architecture with KD results}
\end{table}

As shown in Table~\ref{tb13_THE MLMEC ARCHITECTURE} and Table~\ref{tb14_The MLMEC architecture with KD results}, we compare the amount of data with and without KD using two datasets.
Using the FallAllD dataset, the data amount for the CC was reduced by 34\%, while that for MEC 2 was reduced by 61\%. In the Sisfall dataset, the data amount for the CC was reduced by 71\%, while that for MEC 2 was reduced by 48\%.
In addition, the latency of the MLMEC architecture is calculated by Eq.~\ref{eq25:Ln}.
By comparing Table~\ref{tb13_THE MLMEC ARCHITECTURE} and Table~\ref{tb14_The MLMEC architecture with KD results}, the latency of the MLMEC with KD exhibits a significant reduction compared to that without KD. 
Specifically, latency was remarkably reduced by 54.15\% for the FallAllD dataset and by 46.67\% for the SisFall dataset. 
This indicates that if falls are detected at MEC 1 or MEC 2, there is no need to upload the data to the CC for further detection, thereby reducing computational costs and improving efficiency.

\subsection{Min-Max Normalization and Z-Score Standardization}

N. A. Choudhury~\textit{et al.}~\cite{Review1_Add1} and H. Henderi~\textit{et al.}~\cite{Review1_Add5} compared the detection performance by using min-max normalization and z-score standardization. In~\cite{Review1_Add1}, the obtained test results using min-max normalization and z-score standardization have similar performance in terms of accuracy rate. In~\cite{Review1_Add5}, the obtained test results using min-max normalization outperform z-score standardization in terms of accuracy rate. Based on the testing result that shows min-max normalization outperforms z-score standardization in terms of $ACC$, $REC$, $PRE$, and $F$1 score as shown in Tab.~\ref{tb4_COMPARISON normalization}, we chose the min-max normalization in our experiments.

\begin{table}[htbp]
\centering
\caption{Comparison of min-max normalization and z-score standardization.}
\scalebox{0.9}{
      \begin{tabular}{|c|c|c|c|c|c|}
      \hline
        \textbf{Methods} & \textbf{DL Models} &\textbf{$ACC_{imp}$}&\textbf{$REC_{imp}$}&\textbf{$PRC_{imp}$}&\textbf{$F1_{imp}$}\\
        \hline
        \textbf{\multirow{3}{*}{Min-max}}  & ResNet18 & 82.64\% & 67.94\% &41.61\% &50.85\% \\\cline{2-6}
        & MobileNetV3 & 81.64\% & 68.42\% & 28.37\% & 39.99\% \\\cline{2-6}
        & CNN &  80.44\% & 59.76\% & 31.08\% & 40.39\% \\\hline
        \textbf{\multirow{3}{*}{Z-score}}  & ResNet18 & 76.94\% & 59.38\% &32.31\% & 43.84\% \\\cline{2-6}
        & MobileNetV3 & 62.71\% & 47.23\% & 23.78\% & 29.61\% \\\cline{2-6}
        & CNN &  74.83\% & 55.14\% & 36.28\% & 45.37\% \\\hline   
        \end{tabular}}
     \label{tb4_COMPARISON normalization}
\end{table}

\subsection{Discussions}
We summarized the insights from our experiments as follows:
Without the KD approach, the ACC performances of the ResNet, MobileNetV3, and CNN models are $82.64\%$-$88.12\%$, $67.57\%$-$81.53\%$, and $72.33$-$80.59\%$, respectively.
With the KD approach, the improved ACC performances of combined ResNet, MobileNetV3, and CNN models in the proposed triple-layer MECs are $85.57\%$-$93.89\%$, $85.92\%$-$92.88\%$, and $82.99\%$, respectively.
Meanwhile, the FLOPs of ResNet, MobileNetV3, and CNN in the proposed triple-layer MECs are maintained at 1.292G, 85.87M, and 18.17K, which are the same as before combining, respectively.

The generalizability of the proposed MLMEC FD system with the KD approach is observed as follows:
\begin{itemize}
    \item \emph{Suitability of the Datasets}: FallAllD and SisFall datasets are the sensor data that are collected from triaxial accelerometers. Thus, any sensor datasets collected from triaxial accelerometers can be used to train the proposed MLMEC FD system directly.
    \item \emph{Combinations of DL Models}: Comparing different deployed DL model combinations based on FallAllD and SisFall datasets, the proposed triple-layer MLMEC architecture that adopts ResNet18, MobileNetV3, and CNN in the teacher, TA, and student models is the optimal combination in terms of accuracy rate and latency rate.
    \item \emph{Effectiveness of KD}: For FallAllD and SisFall datasets, the performances of FD are improved significantly by the KD approach. The improved accuracy rates by the KD approach are $2.78\%$ and $11.65\%$ for the FallAllD and SisFall datasets, respectively. The reduced latency rates by the KD approach are $54.15\%$ and $46.67\%$ for the FallAllD and SisFall datasets, respectively.
\end{itemize}

The limitations of the proposed MLMEC FD system with the KD approach are as follows:
\begin{itemize}
    \item \emph{Deployment of DL Models}: The teacher model must have more powerful computational capability than that of the student model. Otherwise, the performance of the student model is degraded by using the smaller DL neural network in the teacher model.
    \item \emph{Computational Capability of MEC Hardware}: the performance of the DL-based model needs to be considered, whether the hardware of the specific MECs can be installed in the real-world application scenarios.
    \item \emph{Types of Application}: The proposed MLMEC architecture is used to solve the classification problem. On the other hand, a regression task may not be appropriately solved by the proposed MLMEC system.
\end{itemize}

\section{Conclusion}
\label{sec:Conclusion}
This study proposes an MLMEC architecture with a KD approach to enhance accuracy and reduce latency.  The FD detection points were divided into multiple devices (EDs, MECs, and CC) to manage the tradeoff between accuracy and latency.  Comparing the performances of the dual- and triple-layer MECs with and without the KD approach, simulation results indicate that triple-layer MECs with the KD approach outperform dual-layer MECs across various confusion matrix metrics. We identified the optimal DL model combination for the triple-layer MEC architecture as ResNet18(teacher), MobileNetV3(TA), and CNN(student). Additionally, a threshold-based judgment mechanism was introduced to distinguish uncertain data and quantify detection levels.  The performance results demonstrate increased certainty in FD at the front-end MEC, reduced uncertainty during the detection process, and decreased data transmission latency, achieving reductions of 54.15\% for the FallAllD and 46.67\% for the SisFall dataset. 

For future works, we summarize as follows: 1) applying the proposed MLMEC architecture to other similar physiological signal recognition, such as electroencephalogram (EEG) and electrocardiography (ECG); 2) implementing the proposed MLMEC framework on the hardware to verify the feasibility; 3) developing more efficient and lightweight models for different layers in the proposed MLMEC framework.

\begin{IEEEbiography}[{\includegraphics[width=1.3in,height=1.7in,clip,keepaspectratio]{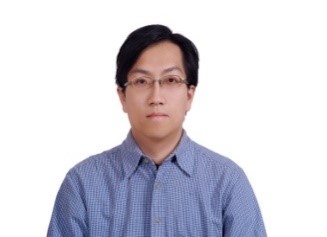}}]{Wei–Lung Mao}was born in Taiwan, in 1972. He received the B.S. degree in electrical engineering from the National Taiwan University of Science and Technology (NTUST), Taipei, Taiwan, in 1994, and the M.S. and Ph.D. degrees in electrical engineering from National Taiwan University (NTU), Taipei, in 1996 and 2004, respectively. 

He is currently a Professor at the Department of Electrical Engineering and the Graduate School of Engineering Science and Technology, National Yunlin University of Science and Technology (NYUST), Douliu, Yunlin, Taiwan. His research interests include satellite navigation systems, intelligent and adaptive control systems, adaptive signal processing, neural networks, and precision control.
\end{IEEEbiography}

\begin{IEEEbiography}[{\includegraphics[width=1in,height=1.25in,clip,keepaspectratio]{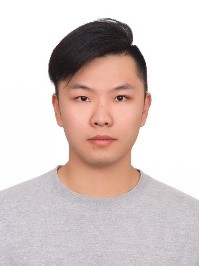}}]{Chun-Chi Wang}was born in Taiwan, in 1995. He received the B.A. degree from the Department of Materials and Energy Engineering, MingDao University, Changhua, Taiwan, in 2018, and the M.S. degree from the Graduate Institute of Aeronautical and Electronic Technology, National Formosa University (NFU), Huwei, Yunlin, Taiwan, in 2021. He is currently pursuing the Ph.D. degree at the National Yunlin University of Science and Technology (NYUST), Douliu, Yunlin, Taiwan. 

His research interests include automated image inspection, deep learning, robotic arm control systems, intelligent manufacturing production lines, and EtherCAT.
\end{IEEEbiography}
 
\begin{IEEEbiography}[{\includegraphics[width=1in,height=1.25in,clip,keepaspectratio]{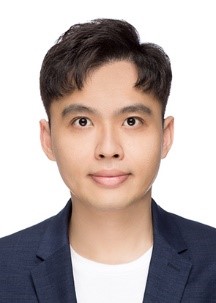}}]{Po-Heng Chou}(Member, IEEE) was born in Tainan, Taiwan. He received the B.S. degree in electronic engineering from National Formosa University (NFU), Huwei, Yunlin, Taiwan, in 2009, the M.S. degree in communications engineering from National Sun Yatsen University (NSYSU), Kaohsiung, Taiwan, in 2011, and the Ph.D. degree from the Graduate Institute of Communication Engineering (GICE), National Taiwan University (NTU), Taipei, Taiwan, in 2020. His research interests include AI for communications, deep learning-based signal processing, wireless networks, and wireless communications, etc.

He was a Postdoctoral Fellow at the Research Center for Information Technology Innovation (CITI), Academia Sinica, Taipei, Taiwan, from Sept. 2020 to Sept. 2024. 
He was a Postdoctoral Fellow at the Department of Electronics and Electrical Engineering, National Yang Ming Chiao Tung University (NYCU), Hsinchu, Taiwan, from Oct. to Dec. 2024.
He has been elected as the Distinguished Postdoctoral Scholar of CITI by Academia Sinica from Jan. 2022 to Dec. 2023. He is invited to visit Virginia Tech (VT) Research Center (D.C. area), Arlington, VA, USA, as a Visiting Fellow, from Aug. 2023 to Feb. 2024.
He received the Partnership Program for the Connection to the Top Labs in the World (Dragon Gate Program) from the National Science and Technology Council (NSTC) of Taiwan to perform advanced research at VT Institute for Advanced Computing (D.C. area), Alexandria, VA, USA, from Jan. 2025 to present.

Additionally, Dr. Chou received the Outstanding University Youth Award and the Phi Tau Phi Honorary Membership from NTU in 2019 to honor his impressive academic achievement. He received the Ph.D. Scholarships from the Chung Hwa Rotary Educational Foundation from 2019 to 2020.
\end{IEEEbiography}

\begin{IEEEbiography}[{\includegraphics[width=1in,height=1.25in,clip,keepaspectratio]{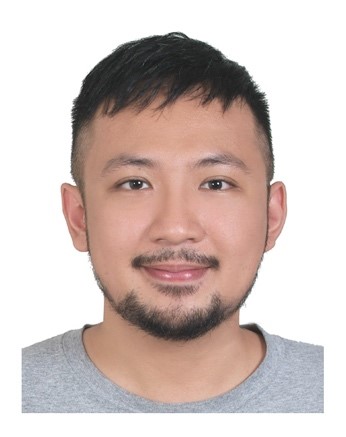}}]{Kai-Chun Liu}(Member, IEEE) received the M.S. and Ph.D. degrees in biomedical engineering from National Yang-Ming University, Taipei, Taiwan, in 2015 and 2019, respectively. 

From 2020 to 2023, he was a Postdoctoral Scholar with the Research Center for Information Technology Innovation, Academia Sinica, Taipei, Taiwan. Currently, he is a Postdoctoral Research Associate with the College of Information and Computer Sciences, University of Massachusetts Amherst, MA, USA. His research interests include pervasive healthcare, wearable computing, machine learning, and biosignal processing.
\end{IEEEbiography}

\begin{IEEEbiography}[{\includegraphics[width=1in,height=1.25in,clip,keepaspectratio]{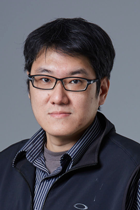}}]{Yu Tsao}(Senior Member, IEEE) received his B.S. and M.S. degrees in Electrical Engineering from National Taiwan University, Taipei, Taiwan, in 1999 and 2001, respectively, and his Ph.D. degree in Electrical and Computer Engineering from the Georgia Institute of Technology, Atlanta, GA, USA, in 2008. 

From 2009 to 2011, he was a researcher at the National Institute of Information and Communications Technology, Kyoto, Japan, where he worked on research and product development in automatic speech recognition for multilingual speech-to-speech translation. He is currently a Research Fellow (Professor) and the Deputy Director of the Research Center for Information Technology Innovation at Academia Sinica, Taipei, Taiwan. He also serves as a Jointly Appointed Professor in the Department of Electrical Engineering at Chung Yuan Christian University, Taoyuan, Taiwan. His research interests include assistive oral communication technologies, audio coding, and bio-signal processing. 

Dr. Tsao is currently an Associate Editor for the IEEE/ACM TRANSACTIONS ON AUDIO, SPEECH, AND LANGUAGE PROCESSING and IEEE SIGNAL PROCESSING LETTERS. He was the recipient of the Academia Sinica Career Development Award in 2017, National Innovation Awards from 2018 to 2021, the Future Tech Breakthrough Award in 2019, the Outstanding Elite Award from the Chung Hwa Rotary Educational Foundation in 2019–2020, the NSTC FutureTech Award in 2022, and the NSTC Outstanding Research Award in 2023. He is the corresponding author of a paper that received the 2021 IEEE Signal Processing Society (SPS) Young Author Best Paper Award.
\end{IEEEbiography}
%\end{comment}
\end{document}